\documentclass[10pt,twocolumn,letterpaper]{article}
\usepackage{caption}
\usepackage{multirow}
\usepackage{cvpr}
\usepackage{times}
\usepackage{epsfig}
\usepackage{subfigure}
\usepackage{graphicx}
\usepackage{amsmath}
\usepackage{amssymb}
\usepackage{float}
\usepackage{tabularx}
\usepackage{authblk}
\usepackage{marvosym}
\usepackage{lipsum}

\usepackage{graphics}
\usepackage{xcolor}
\usepackage{mathrsfs}

\newcommand{\myparagraph}[1]{\vspace{0.1em}\noindent\textbf{#1}}

\newcommand{\zt}[1]{{#1}}

\newcommand{\executeiffilenewer}[3]{%
	\ifnum\pdfstrcmp{\pdffilemoddate{#1}}%
	{\pdffilemoddate{#2}}>0%
	{\immediate\write18{#3}}\fi%
}

\newcommand\blfootnote[1]{%
	\begingroup
	\renewcommand\thefootnote{}\footnote{#1}%
	\addtocounter{footnote}{-1}%
	\endgroup
}

\graphicspath{{./}{inkscape/}{figs/}} 

\usepackage[pagebackref=true,breaklinks=true,letterpaper=true,colorlinks,bookmarks=false]{hyperref}

\cvprfinalcopy 

\def\cvprPaperID{727} 

\ifcvprfinal\fi

\makeatletter 
\renewcommand\AB@affilsepx{\hskip 25pt \protect\Affilfont}
\makeatother

\begin{document}
	\title{OccuSeg: Occupancy-aware 3D Instance Segmentation}
	
	\author[1,2]{Lei Han}
	\author[1]{Tian Zheng}
	\author[1,2]{Lan Xu}
	\author[1\Letter]{Lu Fang}
	\affil[1]{Tsinghua University}
	\affil[2]{Hong Kong University of Science and Technology}
	
	\twocolumn[{%
		\renewcommand\twocolumn[1][]{#1}%
		\maketitle
		\thispagestyle{empty}
		
		\begin{center}
			\centering
			\vspace{-10pt}
			\def\svgwidth{0.9\textwidth}
	\executeiffilenewer{inkscape/teaser.svg}{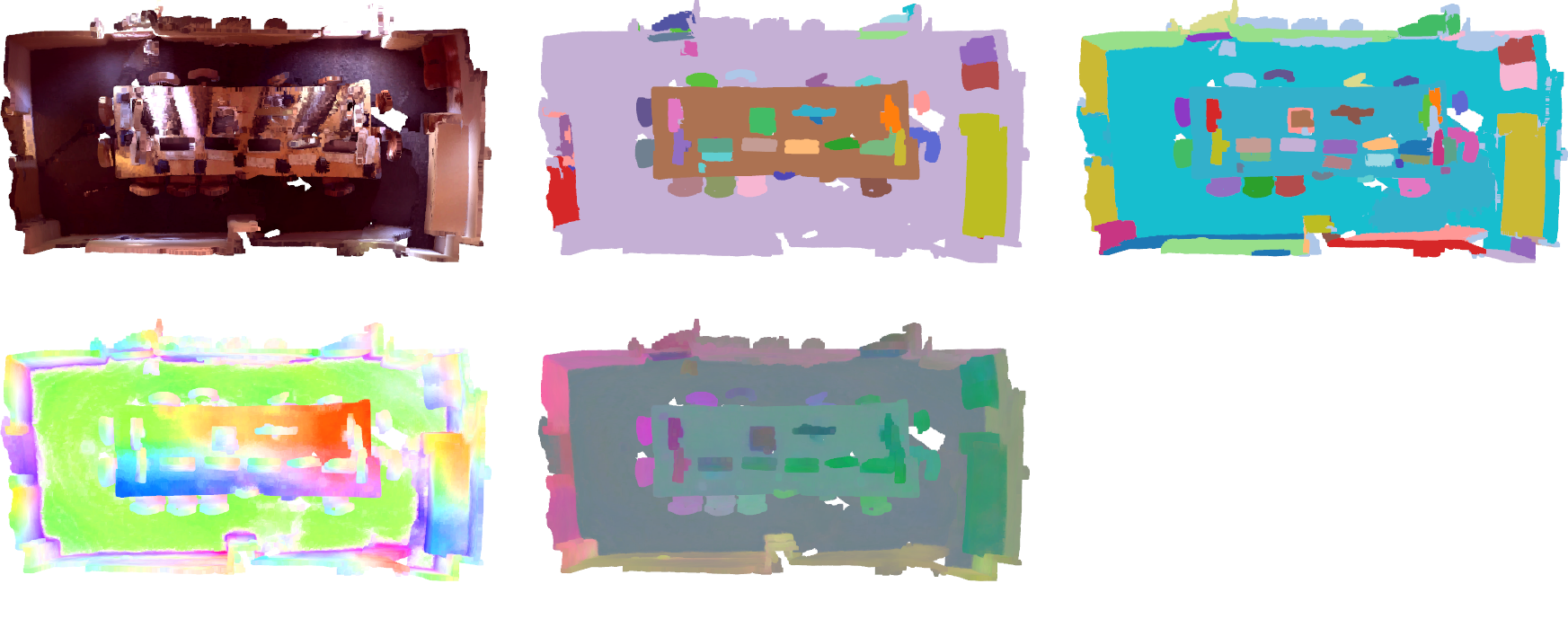}%
	{inkscape -z -D --file=inkscape/teaser.svg %
		--export-pdf=inkscape/teaser.pdf --export-latex}%
\begingroup%
  \makeatletter%
  \providecommand\color[2][]{%
    \errmessage{(Inkscape) Color is used for the text in Inkscape, but the package 'color.sty' is not loaded}%
    \renewcommand\color[2][]{}%
  }%
  \providecommand\transparent[1]{%
    \errmessage{(Inkscape) Transparency is used (non-zero) for the text in Inkscape, but the package 'transparent.sty' is not loaded}%
    \renewcommand\transparent[1]{}%
  }%
  \providecommand\rotatebox[2]{#2}%
  \newcommand*\fsize{\dimexpr\f@size pt\relax}%
  \newcommand*\lineheight[1]{\fontsize{\fsize}{#1\fsize}\selectfont}%
  \ifx\svgwidth\undefined%
    \setlength{\unitlength}{502.49987745bp}%
    \ifx\svgscale\undefined%
      \relax%
    \else%
      \setlength{\unitlength}{\unitlength * \real{\svgscale}}%
    \fi%
  \else%
    \setlength{\unitlength}{\svgwidth}%
  \fi%
  \global\let\svgwidth\undefined%
  \global\let\svgscale\undefined%
  \makeatother%
  \begin{picture}(1,0.39449682)%
    \lineheight{1}%
    \setlength\tabcolsep{0pt}%
    \put(0,0){\includegraphics[width=\unitlength,page=1]{teaser.pdf}}%
    \put(0.15785275,0.20820476){\color[rgb]{0,0,0}\makebox(0,0)[t]{\lineheight{1.25}\smash{\begin{tabular}[t]{c}\textbf{(a) Input Geometry}\end{tabular}}}}%
    \put(0.50113633,0.00376691){\color[rgb]{0,0,0}\makebox(0,0)[t]{\lineheight{1.25}\smash{\begin{tabular}[t]{c}\textbf{(e) Feature Term}\end{tabular}}}}%
    \put(0.15785273,0.00376691){\color[rgb]{0,0,0}\makebox(0,0)[t]{\lineheight{1.25}\smash{\begin{tabular}[t]{c}\textbf{(d) Spatial Term}\end{tabular}}}}%
    \put(0.84441981,0.00376691){\color[rgb]{0,0,0}\makebox(0,0)[t]{\lineheight{1.25}\smash{\begin{tabular}[t]{c}\textbf{(f) Occupancy}\end{tabular}}}}%
    \put(0.50113629,0.20820476){\color[rgb]{0,0,0}\makebox(0,0)[t]{\lineheight{1.25}\smash{\begin{tabular}[t]{c}\textbf{(b) Result}\end{tabular}}}}%
    \put(0.84441981,0.20820476){\color[rgb]{0,0,0}\makebox(0,0)[t]{\lineheight{1.25}\smash{\begin{tabular}[t]{c}\textbf{(c) Ground Truth Instance}\end{tabular}}}}%
    \put(0,0){\includegraphics[width=\unitlength,page=2]{teaser.pdf}}%
  \end{picture}%
\endgroup%

			\vspace{8pt}
			
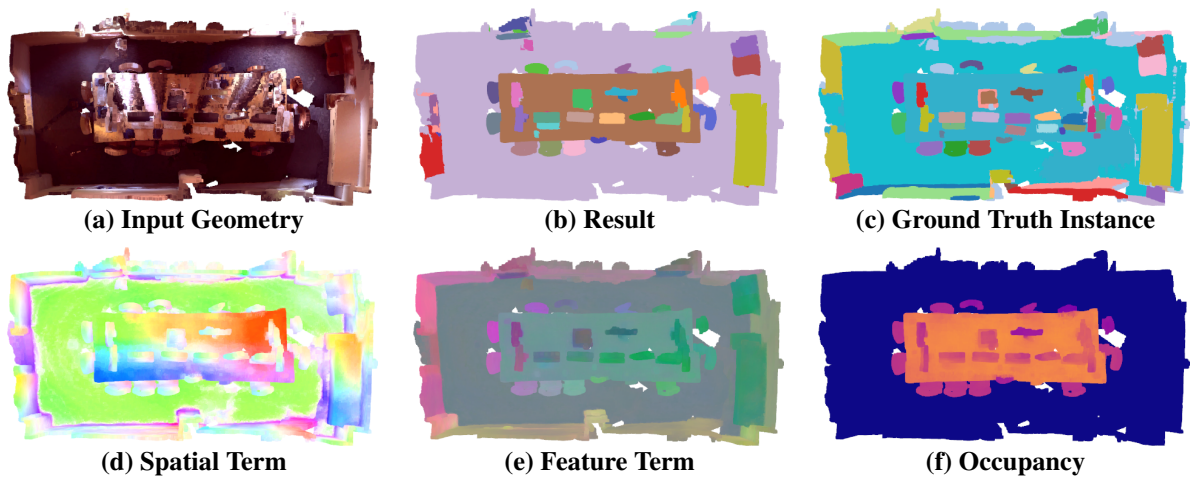
\captionof{figure}{Given the input colored point cloud, occupancy size is regressed for each voxel, which predicts the number of voxels occupied by its belonging instance. An adaptive clustering scheme jointly considers both the occupancy information and embedding distance is further applied for 3D instance segmentation.}
			\label{fig:fig_1_teaser}
		\end{center}
	}]
	
	\blfootnote{\textsuperscript{\Letter} Corresponding author. Mail:  fanglu@sz.tsinghua.edu.cn. \newline \hspace*{1.3em}This work was supported in part by Natural Science Foundation of China (NSFC) under contract No. 61722209 and 6181001011, and was carried out at Tsinghua University.}
	
	\begin{abstract}
		\vspace{-7pt}
		3D instance segmentation, with a variety of applications in robotics and augmented reality, is in large demands these days. 
		Unlike 2D images that are projective observations of the environment, 3D models provide metric reconstruction of the scenes without occlusion or scale ambiguity. 
		In this paper, we define ``3D occupancy size'', as the number of voxels occupied by each instance. It owns advantages of robustness in prediction, on which basis, OccuSeg, an occupancy-aware 3D instance segmentation scheme is proposed. 
		Our multi-task learning produces both occupancy signal and embedding representations, 
		where the training of spatial and feature embedding varies with their difference in scale-aware. Our clustering scheme benefits from the reliable comparison between the {predicted} occupancy size and the {clustered} occupancy size, which encourages hard samples being correctly clustered and avoids over segmentation. The proposed approach achieves state-of-the-art performance on 3 real-world datasets, i.e. ScanNetV2, S3DIS and SceneNN, while maintaining high efficiency.

		
	\end{abstract}
	
	
	\begin{figure*}[ht]
		\centering
		\def\svgwidth{\textwidth}
	\executeiffilenewer{inkscape/pipeline_final.svg}{inkscape/pipeline_final.pdf}%
	{inkscape -z -D --file=inkscape/pipeline_final.svg %
		--export-pdf=inkscape/pipeline_final.pdf --export-latex}%
	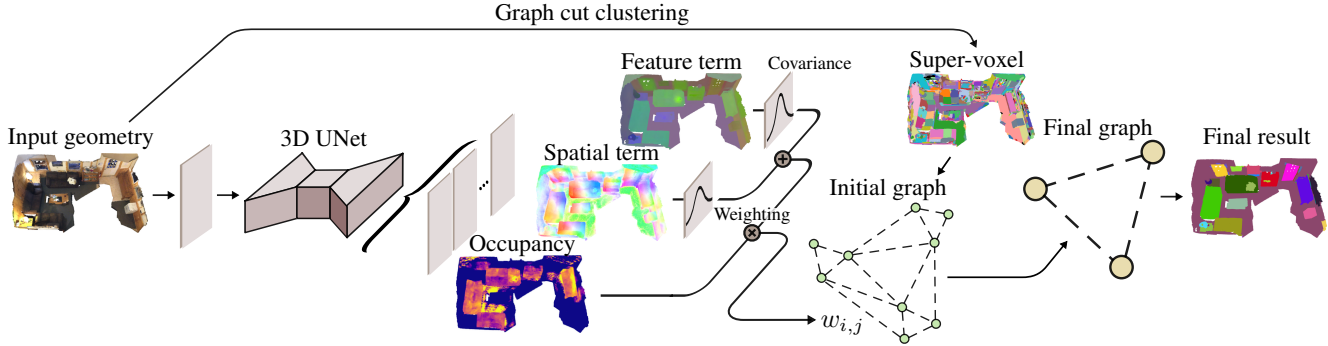%

		\vspace{-10pt}
		\caption{Overview of the proposed instance segmentation scheme. For the input point cloud, our method takes RGB feature as input and employ 3D UNet for point-wise feature learning. The learned feature is decoded to various representations though a fully connected layer for 3D instance segmentation.}
		\vspace{-10pt}
		\label{fig:Network_Overview}
	\end{figure*}
	
	\vspace{-11pt}
	\section{Introduction}
	%
	%
	
	The past ten years have witnessed a rapid development of real-time 3D reconstruction technologies \cite{newcombe2011kinectfusion,niessner2013real,dai2017bundlefusion,LanXu2019,han2018flashfusion} with the popularity of commercial RGB-D depth sensors like Kinect, Xtion, etc.
	Given the reconstructed scene, there is an increasing attention for instance-level semantic understanding of the 3D environment.
	More specifically, 3D instance segmentation aims to recognize points belonging to the same object and simultaneously infer their semantic class, which serves as the fundamental technique for mobile robots as well as augmented/virtual reality applications.
	
	Although scene understanding on 2D images has achieved significant progress recently with the development of deep learning techniques, the irregularity of 3D data introduces new challenges beyond the capability of 2D solutions.
	As demonstrated in previous works~\cite{hou20193d}, directly projecting the state-of-the-art 2D instance segmentation MaskRCNN~\cite{he2017mask} predictions into 3D space leads to poor performance, which inspires better solutions by incorporating 3D geometry information into the network design.
	A popular solution for 3D instance segmentation~\cite{liu2019masc,wang2018sgpn,lahoud20193d} is to marry the powerful 3D feature extractors (spatially sparse convolutional networks~\cite{graham20183d} or PointNet++~\cite{qi2017pointnet2}) with conventional 2D image instance segmentation techniques~\cite{he2017mask,de2017semantic,liu2018affinity}. 
	{Such existing 3D solutions pay less attention to utilizing the inherent property of the 3D model itself, which provides metric reconstruction of the environment without occlusion or scale ambiguity.}
	
	In this paper, we propose an occupancy-aware 3D instance segmentation approach, OccuSeg. It takes the 3D geometry model as input, and produces point-wise predictions of instance level semantic information, as illustrated in Fig. \ref{fig:fig_1_teaser}. 
	Given the insight that the 3D metric space provides more reliable perception than the 2D image-based projective observations for the 3D scene, we particularly introduce ``3D occupancy signal'', representing \textit{the number of voxels occupied by each instance}.
	Such an occupancy signal represents the inherent and fundamental property of each 3D instance, showing a strong potential to handle the ambiguity of scale, location, texture, lighting and occlusion under the 3D setting.
	%
	Thus, we encode the novel occupancy signal into the conventional 3D instance segmentation pipeline, i.e., \textit{learning} stage followed by \textit{clustering} stage.
	In our occupancy-aware approach, both the learning and clustering stages fully utilize the characteristic of the occupancy signal, leading to competitive performance on public datasets.
	The considerable gain on mAP (around 12.3 in mAP) further demonstrates that our occupancy-aware approach owns the superiority of preserving the inherent and fundamental nature of the instances in the 3D environment.
	
	More specifically, the \textit{learning} stage takes a colored 3D scene as input, and utilizes the spatially sparse convolution approaches~\cite{graham2014spatially} to extract a hybrid vector for each voxel~\cite{liu2019masc,liang20193d,lahoud20193d}. \zt{It not only learns the classic embedding such as spatial (Fig. \ref{fig:fig_1_teaser}(d)) and feature embedding (Fig. \ref{fig:fig_1_teaser}(e)),} but also produces an occupancy signal (Fig. \ref{fig:fig_1_teaser}(f)) that implies the object-level volume. \zt{To make full use of both the semantic and geometric information, our feature and spatial embedding are explicitly supervised with different objectives, and are further combined through covariance estimation for both feature and spatial embedding distance.}
	For the \textit{clustering} stage, the 3D input point cloud is grouped into super-voxels based on the geometric and appearance constraints using a graph-based segmentation algorithm~\cite{felzenszwalb2004efficient}. Then, to merge the super-voxels with similar feature embedding into the same instance, we utilize an adaptive threshold to evaluate the similarity between the embedding distance and the occupancy size. Aided by the reliable comparison between the predicted occupancy size and the clustered occupancy size, our clustering encourages hard samples to be correctly clustered and eliminates the false positives where partial instances are recognized as an independent instance. The {technical contributions} are summarized as follows.
	
	
	%
	

	
	\begin{itemize} 
		\setlength\itemsep{0em}
		\item We present an occupancy-aware 3D instance segmentation scheme OccuSeg. It achieves state-of-the-art performance on three public datasets: ScanNetV2~\cite{dai2017scannet}, S3DIS~\cite{armeni20163d} and SceneNN~\cite{hua2016scenenn}, ranking first in all metrics with a significant margin while remaining high efficiency, e.g., $12.3$ gain in mAP on the ScanNetV2 benchmark.
		
		\item In particular, a novel occupancy signal is proposed in this paper, which predicts the number of occupied voxels for each instance. The occupancy signal is learnt jointly with a combination of feature and spatial embedding and employed to guide the clustering stage of 3D instance segmentation.

		
		

	\end{itemize}

	\section{Related Work}
	
	%

	\myparagraph{2D Instance Segmentation.}
	2D Instance segmentation methods are generally divided into two categories: proposal-based and proposal-free approaches.
	Proposal-based methods~\cite{fastRCNN,daiECCV2016_IFCN,he2017mask,BoundaryIS_2017,LiFCN_CVPR2017,xiong2019upsnet} firstly generate region proposals (predefined rectangles) that contain objects and further classify pixels inside each proposal as objects or background.
	By arguing that convolutional operators are translational invariant and thus cannot distinguish similar objects at different places well, Novotny \etal~\cite{novotny2018semi} propose semi-convolutional operators based on the coordinates of each pixel for better instance segmentation.

	On the other hand, proposal-free methods~\cite{liang2017proposal,de2017semantic,Fathi17_dml,Kong_2018_CVPR,leibe2008robust} learn an embedding vector for each pixel and apply a clustering step in the embedding space as post processing for instance segmentation.
	Brabandere~\etal~\cite{de2017semantic} propose to train a per-pixel embedding vector and adopt a discriminative cost to encourage pixels belonging to the same instance to be as close as possible, while the embedding center of different instances to be far away from each other.
	Liang~\etal~\cite{liang2017proposal} regress an offset vector pointing to the object center for each pixel and further use the predicted centers for instance segmentation from a ``voting'' perspective~\cite{leibe2008robust}.
	Recently, Neven~\etal~\cite{neven2019instance} introduce a learnable clustering bandwidth instead of learning embedding using hand-crafted cost functions, achieving accurate instance segmentation in real-time.
	
	While all these approaches have achieved promising results in the 2D domain, 
	the extension to the 3D domain is non-trivial. How to utilize the fundamental property of 3D instance remains a challenging problem.
	
	\myparagraph{3D Instance Segmentation.} 
	Unlike 2D images with regular pixel grids, the irregular distribution of 3D point clouds in physical space raises new challenges for 3D instance segmentation.
	Pioneer works~\cite{wu20153d,tchapmi2017segcloud,hou20193d} have tried to directly extend 2D convolutional neural networks to 3D space by voxelizing the input points into uniform voxels, and applying 3D convolutions instead. Yet most computations are wasted on the inactive empty voxels.
	Thus, recent methods utilize more feasible 3D feature extractors to tackle this problem.
	Point-based instance segmentation approaches \cite{wang2018sgpn,yang2019learning,yi2019gspn} directly consume unordered point clouds as input and use a permutation invariant neural network PointNet~\cite{qi2017pointnet,qi2017pointnet2} for feature extraction. 
	\zt{Landrieu and Simonovsky~\cite{superpoint} generate superpoint graphs from point clouds, followed by ConvGNNs to learn contextual information.}
	While volumetric approaches~\cite{graham2014spatially,choy20194d,vote3deep,liu2019masc,liang20193d,lahoud20193d} take advantage of the sparsity of 3D data and employ sparse convolution to omit computations on inactive voxels. 
	
	Specifically, SGPN~\cite{wang2018sgpn} proposes to learn a similarity matrix for all point pairs, based on which, similar points are merged for instance segmentation.
	3D BoNet~\cite{yang2019learning} directly predicts the bounding boxes of objects for efficient instance segmentation.
	GSPN~\cite{yi2019gspn} introduces a generative shape proposal network and relies on object proposals to identify instances in 3D point clouds.
	\zt{VoteNet~\cite{qi2019deep} predicts offset vectors to the corresponding object centers for seed points, followed by a clustering module to generate object proposals.}
	Additionally, 3DSIS~\cite{hou20193d} jointly learns 2D and 3D features by back projecting features extracted from 2D convolution on images to 3D space. It further applies 3D convolution for volumetric feature learning for proposal-based 3D instance segmentation.
	For proposal-free 3D instance segmentation, MASC~\cite{liu2019masc} combines the SSCN architecture with instance affinity prediction across multiple scales. Liang~\textit{et al.}~\cite{liang20193d} apply embedding learning~\cite{de2017semantic} on top of the superior performance of SSCN.
	Lahoud~\textit{et al.} ~\cite{lahoud20193d} further combine directional information of each object with semantic feature embedding.

	\section{Methods}
	Recall our goal is that, we take a voxelized 3D colored scene as input , and produce a 3D object instance label for each voxel, where the voxels belonging to the same object share an unique instance label. 
	\begin{figure}[t]
		\centering
		\def\svgwidth{\linewidth}
	\executeiffilenewer{inkscape/occupancy_new.svg}{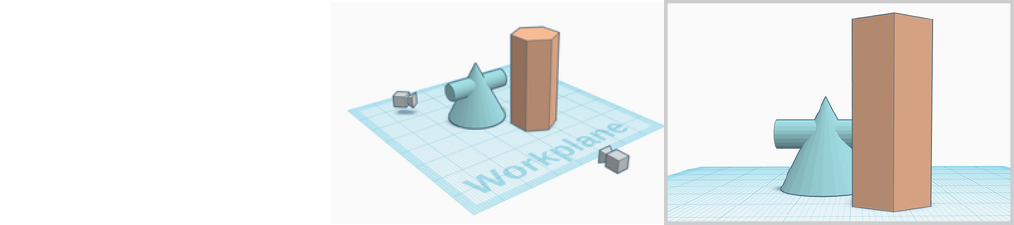}%
	{inkscape -z -D --file=inkscape/occupancy_new.svg %
		--export-pdf=inkscape/occupancy_new.pdf --export-latex}%
\begingroup%
  \makeatletter%
  \providecommand\color[2][]{%
    \errmessage{(Inkscape) Color is used for the text in Inkscape, but the package 'color.sty' is not loaded}%
    \renewcommand\color[2][]{}%
  }%
  \providecommand\transparent[1]{%
    \errmessage{(Inkscape) Transparency is used (non-zero) for the text in Inkscape, but the package 'transparent.sty' is not loaded}%
    \renewcommand\transparent[1]{}%
  }%
  \providecommand\rotatebox[2]{#2}%
  \newcommand*\fsize{\dimexpr\f@size pt\relax}%
  \newcommand*\lineheight[1]{\fontsize{\fsize}{#1\fsize}\selectfont}%
  \ifx\svgwidth\undefined%
    \setlength{\unitlength}{291.9921029bp}%
    \ifx\svgscale\undefined%
      \relax%
    \else%
      \setlength{\unitlength}{\unitlength * \real{\svgscale}}%
    \fi%
  \else%
    \setlength{\unitlength}{\svgwidth}%
  \fi%
  \global\let\svgwidth\undefined%
  \global\let\svgscale\undefined%
  \makeatother%
  \begin{picture}(1,0.22172565)%
    \lineheight{1}%
    \setlength\tabcolsep{0pt}%
    \put(0,0){\includegraphics[width=\unitlength,page=1]{occupancy_new.pdf}}%
    \put(0.73979864,0.1969797){\color[rgb]{0,0,0}\makebox(0,0)[t]{\lineheight{1.25}\smash{\begin{tabular}[t]{c}{\tiny 2D Occupancy:}\end{tabular}}}}%
    \put(0,0){\includegraphics[width=\unitlength,page=2]{occupancy_new.pdf}}%
    \put(0.26021129,0.1969797){\color[rgb]{0,0,0}\makebox(0,0)[t]{\lineheight{1.25}\smash{\begin{tabular}[t]{c}{\tiny 2D Occupancy:}\end{tabular}}}}%
    \put(0,0){\includegraphics[width=\unitlength,page=3]{occupancy_new.pdf}}%
  \end{picture}%
\endgroup%

		\vspace{-6pt}
		\caption{Toy example of 2D observations at different view angles of the same 3D scene. The number of occupied pixels/voxels of each instance (denoted as occupancy) is uncertain on 2D image, yet can be predicted robustly for the reconstructed 3D model.}
		\vspace{-10pt}
		\label{fig:fig_3D2D_visualization}
	\end{figure}
	
	Examining the aforementioned approaches, few of them explicitly utilize the inherent nature of 3D models that differs from 2D image observations: reconstruction of the environment in metric space without occlusion or scale ambiguity. As shown in Fig.~\ref{fig:fig_3D2D_visualization}, for the same instance in 3D space, its observations on 2D images can vary greatly. The number of occupied pixels/voxels of each instance (denoted as occupancy) is unpredictable on 2D image, yet can be predicted robustly from the reconstructed 3D model. 
	
	On the basis of occupancy signal, we propose an occupancy-aware 3D instance segmentation scheme. The pipeline is illustrated in Fig.~\ref{fig:Network_Overview}. While it follows the classic learning followed by clustering procedure, both the learning stage and clustering stage differ from existing approaches.
	First, the input 3D scene is voxelized at a resolution of $2cm$ and is then fed into a 3D convolutional neural network (U-Net~\cite{ronneberger2015u}) for feature extraction.
	Then, the learned feature is forwarded to task-specific heads to learn different representations for each input voxel, including semantic segmentation, which aims to assign a class label, \zt{feature and spatial embedding, as well as occupancy regression} (Sec.~\ref{sec:multi_task_learning}).
	Finally, a graph-based occupancy-aware clustering scheme is performed, which utilizes both the predicted occupancy information and the feature embedding from the previous stage (Sec.~\ref{sec:instance_inference}). 
	Note that all the 3D convolutions are realized using a submanifold sparse convolutional network~\cite{graham20183d} to employ the sparsity nature of input 3D scene.
	The details of the network are provided in the Appendix.
	
	\subsection{Multi-task Learning}
	\label{sec:multi_task_learning}
	In order to jointly leverage the inherent occupancy and semantic and spatial information from the 3D scene, we propose a multi-task learning framework to learn task-specific representations for the $i$-th input voxel, including (1) $\mathbf{c}_i$ for the semantic segmentation, which aims to assign a class label; (2) $\mathbf{s}_i$ and $\mathbf{d}_i$ for the joint feature and spatial embedding, as well as the corresponding $\mathbf{b}_i$ for covariance prediction to fuse feature and spatial information; and (3) ${o}_i$ for the occupancy regression.
	The network is trained to minimize a joint cost function $\mathcal{L}_{\mathrm{joint}}$:
	\begin{equation}
	\mathcal{L}_{\mathrm{joint}} = \mathcal{L}_{\mathrm{c}} + \mathcal{L}_{\mathrm{e}} + \mathcal{L}_{\mathrm{o}}.
	\end{equation}
	Here $\mathcal{L}_{\mathrm{c}}$ is a conventional cross-entropy loss~\cite{goodfellow2016deep} for semantic segmentation.
	$\mathcal{L}_{\mathrm{e}}$ aims to learn an embedding vector that considers jointly feature and spatial embedding for instance segmentation (Sec.~\ref{sec:embedding_learning}).
	$\mathcal{L}_{\mathrm{o}}$ serves for the regression of the occupancy size of each voxel's belonging instance (Sec.~\ref{sec:occupancy_regression}).
	
	\subsubsection{Embedding Learning}
	\label{sec:embedding_learning}
	%
	Unlike previous methods~\cite{novotny2018semi} that concatenate the feature and spatial embedding directly, we propose to separate them explicitly and supervise their learning process with different objectives. 
	Our key observation is that while spatial embedding is scale-aware and has an explicit physical explanation, such as an offset vector from current voxel to the spatial center of its belonging instance, feature embedding suffers from inherently ambiguous scale, and thus has to be regularized using additional cost functions.
	Both embeddings are further regularized using the covariance estimation. 
	Our learning function for embedding $\mathcal{L}_\mathrm{e}$ consists of three terms, i.e., spatial term $\mathcal{L}_\mathrm{sp}$, feature term $\mathcal{L}_\mathrm{se}$, and covariance term $\mathcal{L}_\mathrm{cov}$,  
	\begin{align}
	\mathcal{L}_\mathrm{e} &= \mathcal{L}_\mathrm{sp} +  \mathcal{L}_\mathrm{se} + \mathcal{L}_\mathrm{cov}.
	\end{align}
	

	\myparagraph{Spatial Term.}
	Spatial embedding $\mathbf{d}_i$ for the $i$-th voxel is a 3-dimensional vector that regresses to the object center, which is supervised using the following spatial term:
	\begin{equation}
	\mathcal{L}_{\mathrm{sp}} = \frac{1}{C}\sum_{c=1}^{C}\frac{1}{N_c}\sum_{i=1}^{N_c}||\mathbf{d}_i + \mathbf{\mu}_i- \frac{1}{N_c}\sum_{i=1}^{N_c}\mu_i||,
	\end{equation}
	where $C$ is the number of instances in the input 3D scene, $N_c$ is the number of voxels in the $c$-th instance, and $\mathbf{\mu}_i$ represents the 3D position of the $i$-th voxel of the $c$-th instance. 
	
	\myparagraph{Feature Term.} 
	Feature embedding $\mathbf{s}_i$ is learned using a discriminative loss function~\cite{de2017semantic} that consists of three terms: 
	\begin{equation}
	\mathcal{L}_\mathrm{se} = \mathcal{L}_\mathrm{var} + \mathcal{L}_\mathrm{dist} + \mathcal{L}_\mathrm{reg},
	\end{equation}
	where the variance term $\mathcal{L}_\mathrm{var}$ draws current embedding towards the mean embedding of each instance, the distance term $\mathcal{L}_\mathrm{dist}$ pushes instances away from each other, and the regularization term $\mathcal{L}_\mathrm{reg}$ draws all instances towards the origin to keep the activation bounded. The detailed formulations are as follows.
	\begin{align}
	\mathcal{L}_\mathrm{var}	&= \frac{1}{C}\sum_{c=1}^{C}\frac{1}{N_c}\sum_{i=1}^{N_C}[||\mathbf{u}_c-\mathbf{s}_i||-\delta_v]_{+}^2, \\
	\mathcal{L}_\mathrm{dist} &= \frac{1}{C(C-1)}\sum_{c_A=1}^{C}\sum_{c_B=c_A+1}^{C}[2\delta_d - ||\mathbf{u}_{c_A}-\mathbf{u}_{c_B}||]_{+}^2,	\\
	\mathcal{L}_\mathrm{reg} &= \frac{1}{C}\sum_{c=1}^{C}||\mathbf{u}_c||.
	\end{align}
	Here, $\mathbf{u}_c = \frac{1}{N_c}\sum_{i=1}^{N_c}\mathbf{s}_i$ represents the mean feature embedding of the $c$-th instance. The predefined thresholds $\delta_v$ and $\delta_d$ are set to be {0.1 and 1.5}, ensuring that the intra-instance embedding distance is smaller than the inter-instance distance.
	
	\myparagraph{Covariance Term.} 
	The covariance term aims to learn an optimal clustering region for each instance.
	Let $\mathbf{b}_i = (\sigma_s^i,\sigma_d^i)$ denote the predicted feature/spatial covariance for the $i$-th voxel in the $c$-th instance. By averaging $\mathbf{b}_i$, we obtain $(\sigma_{s}^{c}, \sigma_{d}^{c})$, the embedding covariance of the $c$-th instance.
	Then, the probability of the $i$-th voxel belonging to the $c$-th instance, denoted as $p_i$, is formulated as:
	%
	\begin{equation}
	\label{eqn_probability}
	p_i = \exp(-(\frac{||\mathbf{s}_i - \mathbf{u}_c||}{\sigma_{s}^{c}})^2 - (\frac{||\mathbf{\mu}_i + \mathbf{d}_i - \mathbf{e}_c||}{\sigma_{d}^{c}})^2),
	\end{equation}
	where $\mathbf{e}_c = \frac{1}{N_c}\sum_{k=0}^{N_c}(\mathbf{\mu}_k + \mathbf{d}_k)$ represents the predicted spatial center of the $c$-th instance.
	%
	%
	Since $p_i$ is expected to be larger than $0.5$ for voxels that belong to the $c$-th instance, the covariance term is then formulated by a binary cross-entropy loss,
	\begin{equation}
	\mathcal{L}_\mathrm{cov} = -\frac{1}{C}\sum_{c=1}^{C}\frac{1}{N}\sum_{i=1}^{N}[y_ilog(p_i)+(1-y_i)log(1-p_i)],
	\end{equation}
	where $y_i = 1$ indicates $i$ belongs to $c$ and $y_i=0$ otherwise, $N$ indicates the number of points in the input point cloud.

	\subsubsection{Occupancy Regression}
	\label{sec:occupancy_regression}
	To utilize the occupancy information under the 3D setting, for the $i$-th voxel in the $c$-th instance, we predict a positive value $o_i$ to indicate the number of voxels occupied by the current instance.
	Then, the average of $o_i$ will serve as the predicted occupancy size of the current instance.
	%
	For more robust prediction, we regress the logarithm instead of the original value and formulate the following occupancy term,
	\begin{equation}
	\mathcal{L}_\mathrm{o} = \frac{1}{C}\sum_{c=1}^{C}\frac{1}{N_c}\sum_{i=1}^{N_c}||o_i - \log(N_c)||,
	\end{equation}
	where $N_c$ is the number of voxels in the $c$-th instance.

\begin{figure}[tbp]
		\begin{center}
			\centering
			\includegraphics[width=8cm]{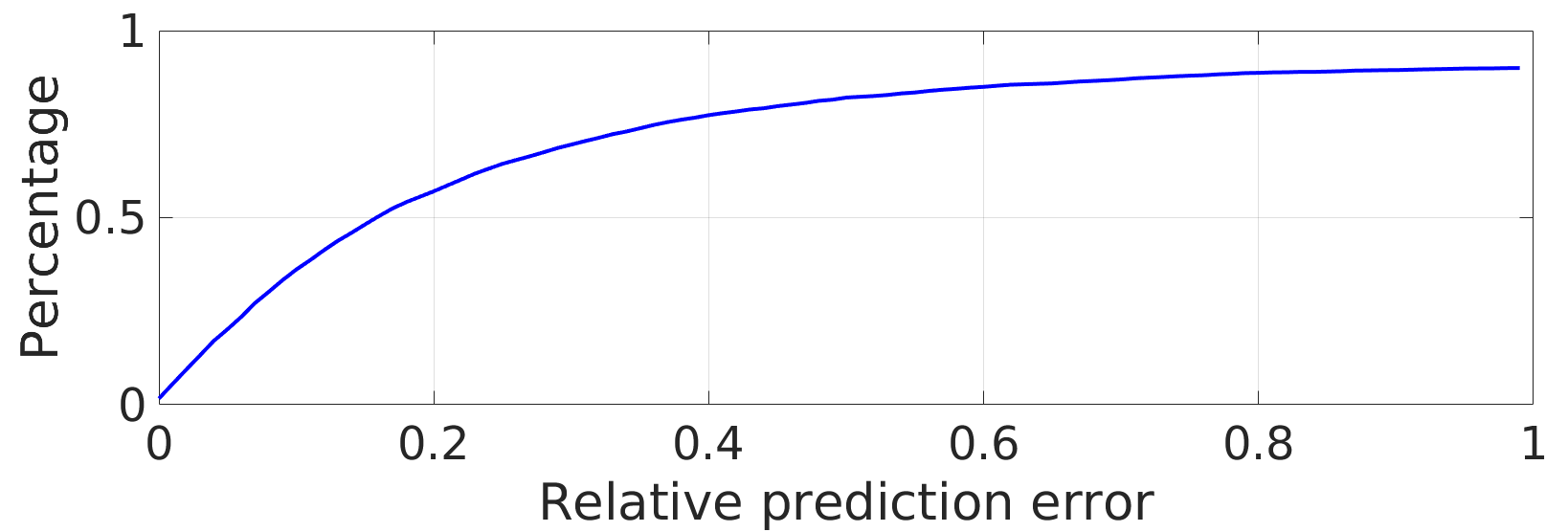}
		\end{center}
		\vspace{-2pt}
		\caption{Cumulative distribution function of the relative prediction error on the validation set of the ScanNetV2~\cite{dai2017scannet}.} 
		\vspace{-8pt}
		\label{fig:occupancy_regression}
	\end{figure}

	To evaluate the feasibility of our occupancy prediction strategy, we use the relative prediction error $R_c$ to measure the occupancy prediction performance of the $c$-th instance,
	\begin{equation}
	R_c = \frac{|N_c - \exp(\frac{1}{N_c}\sum_{i=1}^{N_c}o_i)|}{N_c}.
	\end{equation}
	{We particularly plot the cumulative distribution function of $R_c$ in Fig.~\ref{fig:occupancy_regression}.} For over 4000 instances in the validation set of ScanNetV2 dataset~\cite{dai2017scannet}, more than $68\%$ instances are predicted, with a relative error smaller than $0.3$, which illustrates the effectiveness of our occupancy regression for the following clustering stage.
	
	\begin{figure}[t]
		\centering
		\def\svgwidth{\columnwidth}
	\executeiffilenewer{inkscape/compare2.svg}{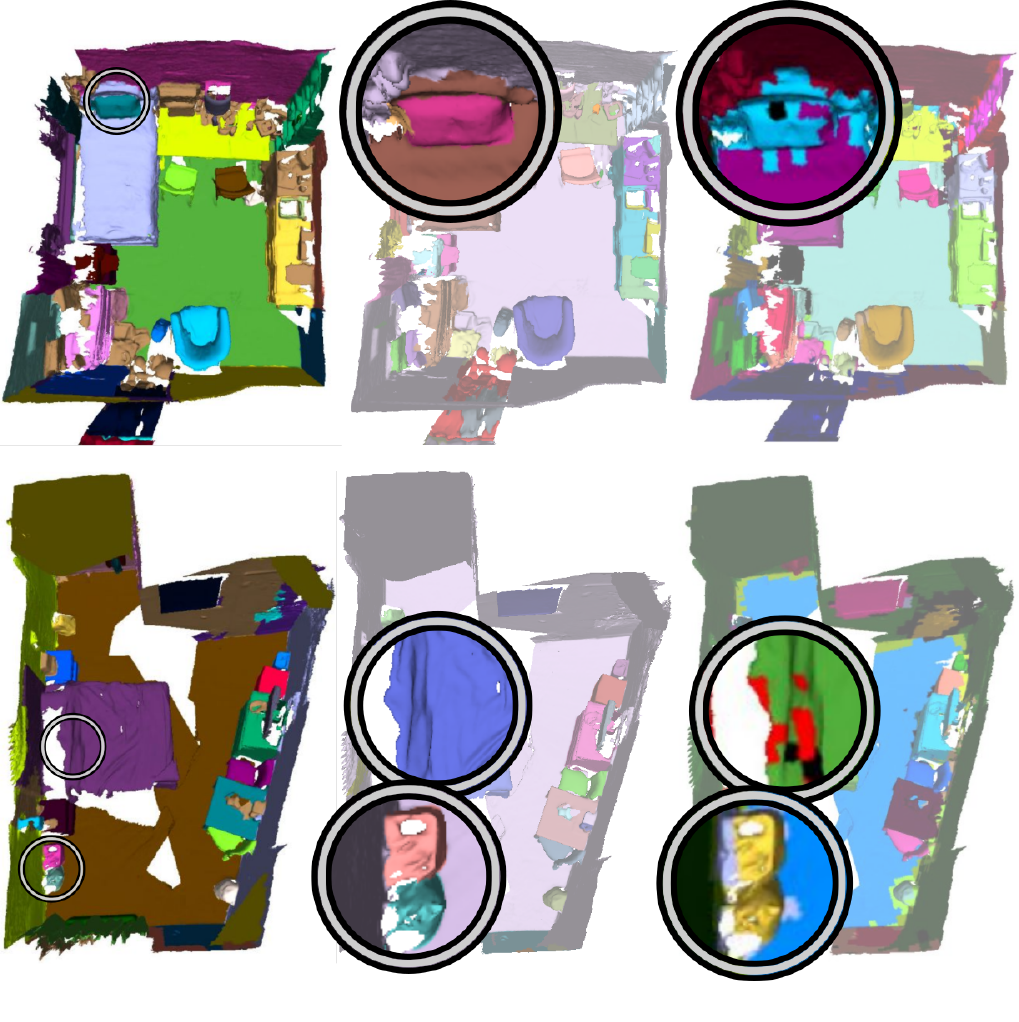}%
	{inkscape -z -D --file=inkscape/compare2.svg %
		--export-pdf=inkscape/compare2.pdf --export-latex}%
\begingroup%
  \makeatletter%
  \providecommand\color[2][]{%
    \errmessage{(Inkscape) Color is used for the text in Inkscape, but the package 'color.sty' is not loaded}%
    \renewcommand\color[2][]{}%
  }%
  \providecommand\transparent[1]{%
    \errmessage{(Inkscape) Transparency is used (non-zero) for the text in Inkscape, but the package 'transparent.sty' is not loaded}%
    \renewcommand\transparent[1]{}%
  }%
  \providecommand\rotatebox[2]{#2}%
  \newcommand*\fsize{\dimexpr\f@size pt\relax}%
  \newcommand*\lineheight[1]{\fontsize{\fsize}{#1\fsize}\selectfont}%
  \ifx\svgwidth\undefined%
    \setlength{\unitlength}{293.28000767bp}%
    \ifx\svgscale\undefined%
      \relax%
    \else%
      \setlength{\unitlength}{\unitlength * \real{\svgscale}}%
    \fi%
  \else%
    \setlength{\unitlength}{\svgwidth}%
  \fi%
  \global\let\svgwidth\undefined%
  \global\let\svgscale\undefined%
  \makeatother%
  \begin{picture}(1,1.00333469)%
    \lineheight{1}%
    \setlength\tabcolsep{0pt}%
    \put(0.14601389,0.00192042){\color[rgb]{0,0,0}\makebox(0,0)[t]{\lineheight{1.25}\smash{\begin{tabular}[t]{c}GT\end{tabular}}}}%
    \put(0.47870745,0.00192042){\color[rgb]{0,0,0}\makebox(0,0)[t]{\lineheight{1.25}\smash{\begin{tabular}[t]{c}Ours\end{tabular}}}}%
    \put(0.82177697,0.00192042){\color[rgb]{0,0,0}\makebox(0,0)[t]{\lineheight{1.25}\smash{\begin{tabular}[t]{c}Lahoud \etal~\cite{lahoud20193d}\end{tabular}}}}%
    \put(0,0){\includegraphics[width=\unitlength,page=1]{compare2.pdf}}%
  \end{picture}%
\endgroup%

		\vspace{2pt}
		\caption{Qualitative comparisons between OccuSeg and a previous approach~\cite{lahoud20193d} on the validation set of the ScanNetV2~\cite{dai2017scannet}. OccuSeg generates more consistent instance labels and successfully distinguishes nearby small instances thanks to the proposed occupancy aware clustering scheme.}
		\vspace{-8pt}
		\label{fig:compare2}
	\end{figure}

	\begin{figure*}[tbp]
		\centering
		\def\svgwidth{\textwidth}
	\executeiffilenewer{inkscape/comparison.svg}{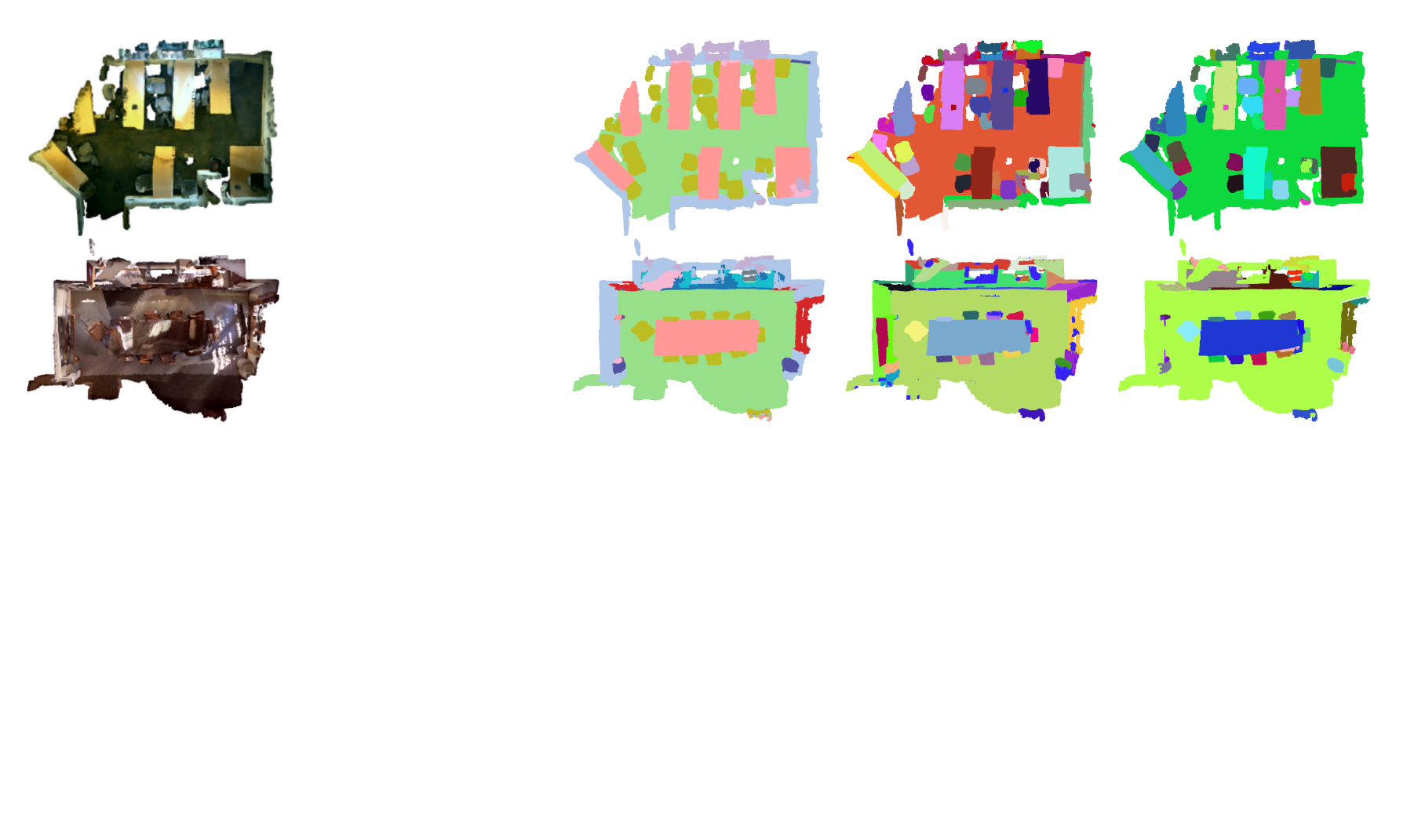}%
	{inkscape -z -D --file=inkscape/comparison.svg %
		--export-pdf=inkscape/comparison.pdf --export-latex}%
\begingroup%
  \makeatletter%
  \providecommand\color[2][]{%
    \errmessage{(Inkscape) Color is used for the text in Inkscape, but the package 'color.sty' is not loaded}%
    \renewcommand\color[2][]{}%
  }%
  \providecommand\transparent[1]{%
    \errmessage{(Inkscape) Transparency is used (non-zero) for the text in Inkscape, but the package 'transparent.sty' is not loaded}%
    \renewcommand\transparent[1]{}%
  }%
  \providecommand\rotatebox[2]{#2}%
  \newcommand*\fsize{\dimexpr\f@size pt\relax}%
  \newcommand*\lineheight[1]{\fontsize{\fsize}{#1\fsize}\selectfont}%
  \ifx\svgwidth\undefined%
    \setlength{\unitlength}{518.96162588bp}%
    \ifx\svgscale\undefined%
      \relax%
    \else%
      \setlength{\unitlength}{\unitlength * \real{\svgscale}}%
    \fi%
  \else%
    \setlength{\unitlength}{\svgwidth}%
  \fi%
  \global\let\svgwidth\undefined%
  \global\let\svgscale\undefined%
  \makeatother%
  \begin{picture}(1,0.59431295)%
    \lineheight{1}%
    \setlength\tabcolsep{0pt}%
    \put(0,0){\includegraphics[width=\unitlength,page=1]{comparison.pdf}}%
    \put(0.10933699,0.58099045){\color[rgb]{0,0,0}\makebox(0,0)[t]{\lineheight{1.25}\smash{\begin{tabular}[t]{c}\textbf{Input Geometry}\end{tabular}}}}%
    \put(0.51194193,0.58099045){\color[rgb]{0,0,0}\makebox(0,0)[t]{\lineheight{1.25}\smash{\begin{tabular}[t]{c}\textbf{Predicted Semantic}\end{tabular}}}}%
    \put(0.70802267,0.58099045){\color[rgb]{0,0,0}\makebox(0,0)[t]{\lineheight{1.25}\smash{\begin{tabular}[t]{c}\textbf{GT Instance}\end{tabular}}}}%
    \put(0.88080813,0.58099045){\color[rgb]{0,0,0}\makebox(0,0)[t]{\lineheight{1.25}\smash{\begin{tabular}[t]{c}\textbf{Predicted Instance}\end{tabular}}}}%
    \put(0,0){\includegraphics[width=\unitlength,page=2]{comparison.pdf}}%
    \put(0.30809254,0.58099045){\color[rgb]{0,0,0}\makebox(0,0)[t]{\lineheight{1.25}\smash{\begin{tabular}[t]{c}\textbf{GT Semantic}\end{tabular}}}}%
  \end{picture}%
\endgroup%

		\vspace{-4pt}
		\caption{Representative 3D instance segmentation results on the validation set of public datasets, including ScanNetV2~\cite{dai2017scannet} and S3DIS~\cite{armeni20163d}.}
		\vspace{-10pt}
		\label{fig:all}
	\end{figure*}
	\subsection{Instance Clustering}
	\label{sec:instance_inference}
	
	
	In this subsection, based on the multi-representation learning from the previous stage, a graph-based occupancy-aware clustering scheme is introduced to tackle the 3D instance segmentation problem during inference. 
	Specifically, we adopt a bottom-up strategy and group the input voxels into super-voxels using an efficient graph-based segmentation scheme~\cite{felzenszwalb2004efficient}. 
	Compared with super-pixel representations in 2D space~\cite{schuurmans2018efficient,yang2017semantic}, super-voxel representation works better to separate different instances where the instance boundaries in 3D space is easier to identify thanks to the geometry continuity or local convexity constraints~\cite{christoph2014object}.
	
	Let $\Omega_i$ denote the collection of all the voxels belonging to the super-voxel $v_i$, we define the spatial embedding $\mathbf{D}_i$ of $v_i$ as,
	\begin{equation}
	\mathbf{D}_i = \frac{1}{|\Omega_i|}\sum_{k\in\Omega_i}(\mathbf{d}_i + \mathbf{\mu}_i),\\
	\end{equation}
	where $|\Omega_i|$ represents the number of voxels in $\Omega_i$. 
	The feature embedding $\mathbf{S}_i$, occupancy ${O}_i$ and covariance ${\sigma_s^i},{\sigma_d^i}$ of $v_i$ are computed based on a similar averaging operation for all the voxels belonging to $v_i$. 
	We further define the following occupancy ratio $r_i$ to guide the clustering step,
	\begin{equation}
	r_i = \frac{O_i}{|\Omega_i|}.
	\end{equation}
	{Note that $r_i > 1$ indicates that there are too many voxels in $v_i$ for instance segmentation, otherwise $v_i$ should attract more voxels.}
	
	Given the super-voxel representation, an undirected graph $G = (V, E, W)$ is established, where the vertices $v_i \in V$ represent the generated super-voxels, $e_{i,j} = (v_i, v_j) \in E$ indicates the pairs of vertices with a weight $w_{i,j} \in W$. The weight $w_{i,j}$ represents the similarity between $v_i$ and $v_j$. Here $w_{i,j}$ is formulated as
	\begin{equation}
	\label{eqn_weight}
	w_{i,j} = \frac{\exp(-(\frac{||\mathbf{S_i}-\mathbf{S_j}||}{\sigma_s})^2 - (\frac{||\mathbf{D_i}-\mathbf{D_j}||}{\sigma_d})^2)}{max(r,0.5)},    
	\end{equation}
	where $\sigma_s, \sigma_d$ and $r$ represent the feature covariance, spatial covariance and occupancy ratio of the virtual super-voxel that merges both $v_i$ and $v_j$.

	%
	Note that a larger weight indicates a higher possibility that $v_i$ and $v_j$ belong to the same instance. 
	And during the calculation of the merging weight, our occupancy ratio helps to punish over-segmented instances and encourages partial instances to be merged together as shown in Fig.~\ref{fig:compare2}.

	For all the edges in $E$, we select edge $e_{i,j}$ with the highest weight $w_{i,j}$ and merge $v_i, v_j$ as a new vertex if $w_{i,j} > T_0$, where the merge threshold $T_0$ is set to be $0.5$. 
	The graph $G$ is then updated after every merge operation. 
	This process is iterated until none of the weight is larger than $T_0$. 
	Finally, the remaining vertices in $G$ are labeled as instances if their occupancy ratio $r$ satisfies the constraint of $0.3 < r < 2$ to reject false positives in instance segmentation. 
	%
	

	\subsection{Network Training}
	\label{sec:trainig_details}
	We employ a simple UNet-like structure~\cite{ronneberger2015u} for feature extraction from the input point cloud with color information. Network details are presented in the Appendix. For the sake of efficiency, the chunk-based sparse convolution strategy {in~\cite{han2019Semantic}} is adopted, which is $4\times$ faster than the original implementation of the SCN~\cite{graham20183d}. The network is trained using Adam optimizer with an initial learning rate of 1e-3. For all the datasets including ScanNetV2~\cite{dai2017scannet}, Stanford3D~\cite{armeni20163d} and SceneNN~\cite{hua2016scenenn} as shown in the experiments of Sec.~\ref{sec_exp}, we use the same hyper-parameters and train the network from scratch for 320 epochs.

	\begin{figure*}[tbp]
		\centering
		\def\svgwidth{\textwidth}
	\executeiffilenewer{inkscape/real.svg}{inkscape/real.pdf}%
	{inkscape -z -D --file=inkscape/real.svg %
		--export-pdf=inkscape/real.pdf --export-latex}%
	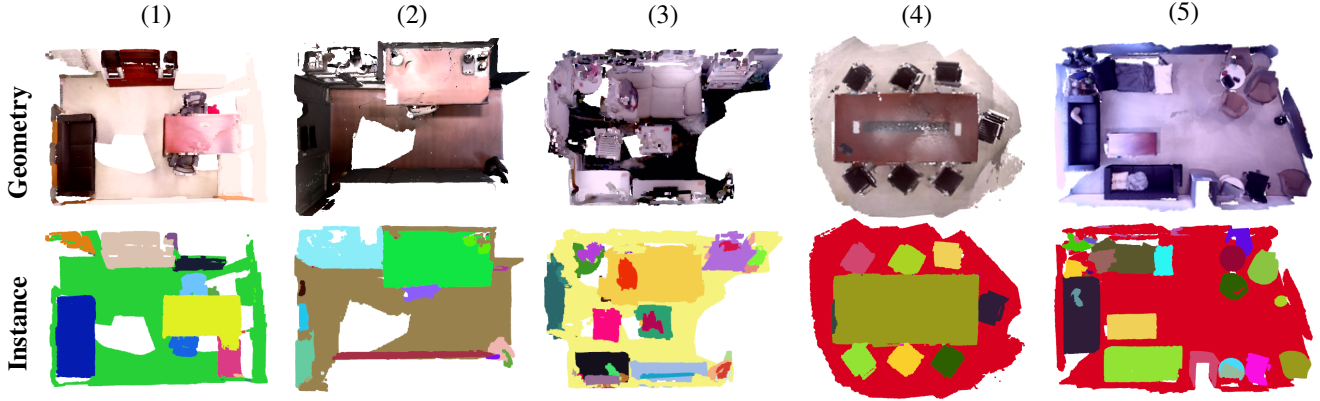%

		\vspace{-10pt}
		\caption{3D instance segmentation results on real-world scenes. Here the 3D geometric models are reconstructed using FlashFusion~\cite{han2018flashfusion} system, with the input being the depth and color sequences from consumer-level RGB-D camera. Our scheme generates robust instance segmentation results in real-world environments using the network trained on a public dataset, ScanNetV2~\cite{dai2017scannet}. }
		\label{fig:fig_flash_instance}
	\end{figure*}

	\begin{table*}[htbp]
		\centering
		\small
		\setlength{\tabcolsep}{0.9mm}{
			\begin{tabular}{|c|c|cccccccccccccccccc|}
				\hline
				Method   & \textbf{mAP} &{bath}  & {bed} & bkshf &cab & {chair}& {cntr}& {curt}  & {desk} & {door}&{ofurn} & pic   & {fridg} & {showr}& {sink} & {sofa} & {tabl}  &{toil} & {wind}     \\
				\hline
				3D-SIS~\cite{hou20193d} & 16.1 & 40.7 &	15.5 &	6.8 &	4.3 &	34.6 & 0.1&	13.4 &	0.5 &	8.8 &	10.6 & 3.7 & 13.5 &	32.1 &	2.8 &	33.9 &	11.6 &	46.6 &	9.3 
				\\
				PanopticFusion~\cite{narita2019panopticfusion} & 21.4 & 25.0 &	33.0 &	27.5 &	10.3 &	22.8 &	0.0 &	34.5 &	2.4 &8.8 &	20.3 &	18.6 &	16.7 &	36.7 &	12.5 &	22.1 &	11.2 &	66.6 &	16.2 
				\\
				3D-BoNet~\cite{yang2019learning} & 25.3  & 51.9 &	32.4 &	25.1 &	13.7 &	34.5 & 3.1 &	41.9 &	6.9 &	16.2 &	13.1 &	5.2 &20.2 &	33.8 &	14.7 &	30.1 &	30.3 &	65.1 & 17.8
				
				\\
				MTML~\cite{lahoud20193d} & 28.2  & 57.7 &	38.0 &	18.2 &	10.7 &	43.0 &	0.1 &	\textbf{42.2} &	5.7 &	17.9 &	16.2 &	7.0 &	22.9 &	51.1 &	16.1 &	49.1 &	31.3 &	65.0 &	16.2
				
				\\
				Occipital-SCS  & 32.0 &  67.9 &	35.2 &	33.4 &	22.9 &	43.6 &	2.5 &	41.2 &	5.8 &	16.1 &	24.0 &	8.5 &	26.2 &	49.6 &	18.7 &	46.7 &	32.8 &	77.5 &	23.1
				
				\\
				OccuSeg  &  \textbf{44.3} &	\textbf{85.2} &	\textbf{56.0} &	\textbf{38.0} & \textbf{24.9} &	\textbf{67.9} &	\textbf{9.7} &	34.5 &	\textbf{18.6} &	\textbf{29.8} &	\textbf{33.9} &	\textbf{23.1} &	\textbf{41.3} &	\textbf{80.7} &	\textbf{34.5} &	\textbf{50.6} &	\textbf{42.4} &	\textbf{97.2} &	\textbf{29.1} 
				\\
				\hline
			\end{tabular}
			\vspace{8pt}
			\caption{ Quantitative comparison on the ScanNetV2~\cite{dai2017scannet} benchmark in terms of mAP score on 18 classes. Our approach achieves \textbf{the best performance in 17 out of 18 classes}. Note that the
				ScanNetV2 benchmark data is accessed on 11/14/2019.}
			\label{tab:perclass_iou_scannet}
		}
	\end{table*}

	\section{Experiments} \label{sec_exp}
	In this section, we evaluate our method on a variety of challenging scenarios.
	%
	For experiments on public datasets, we run our method on a PC with a NVIDIA TITAN Xp GPU and an Intel(R) Xeon(R) E5-2650 CPU. 
	For real-world experiments, our method is conducted on the laptop Microsoft Surface Book 2 with a NVIDIA GTX 1060 (Mobile) GPU and an Intel Core i7-8650U CPU. Using the real-time 3D reconstruction method FlashFusion~\cite{han2018flashfusion} for the 3D geometric input, we present the demo of online 3D instance segmentation on the portable device. More details are provided in the supplementary video.
	
	We employ the popular 3D instance segmentation benchmark ScanNetV2~\cite{dai2017scannet}, as well as the widely used S3DIS~\cite{armeni20163d} and SceneNN~\cite{hua2016scenenn} datasets. 
	ScanNetV2 benchmark~\cite{dai2017scannet} contains 1513 indoor RGBD scans with 3D instance annotations, while Stanford Large-Scale 3D Indoor Space Dataset (S3DIS)~\cite{armeni20163d} contains 6 large-scale indoor areas covering over $6000 m^2$ with 13 object classes.
	{SceneNN}~\cite{hua2016scenenn} is a smaller indoor 3D dataset with 50 scans as the training set and 26 scans for evaluation, which is used to evaluate our performance under less training data.

	\subsection{Qualitative Evaluation}
	
	The representative 3D instance segmentation results on the validation set of public datasets are presented in Fig.~\ref{fig:all}, which demonstrate that the proposed approach achieves robust instance segmentation results for complex environments.

	To further verify the robustness of our method on real-world scenes, we implement our method on the basis of real-time 3D reconstruction method FlashFusion~\cite{han2018flashfusion} for online 3D instance segmentation. 
	As shown in Fig.~\ref{fig:fig_flash_instance}, our network pre-trained on ScanNetV2 can generate 3D instance segmentation results robustly in real-world scenarios.
	More live results are provided in the supplementary video.

	\begin{table}[tbp]
		\centering
		\begin{tabular}{|c|c|c|c|}
			\hline
			& mAP  & mAP@0.5 & mAP@0.25 \\
			\hline
			3D-SIS~\cite{hou20193d} & 16.1 & 38.2 & 55.8 \\
			\hline
			3D-BoNet~\cite{yang2019learning} & 25.3 & 48.8 & 68.7 \\
			\hline
			MASC~\cite{liu2019masc} & 25.4 & 44.7 & 61.5 \\
			\hline
			MTML~\cite{lahoud20193d} & 28.2 & 54.9 & 73.1 \\
			\hline
			Occipital-SCS & 32.0 &	51.2	&	68.8	\\
			\hline
			OccuSeg & \textbf{44.3} & \textbf{63.4} & \textbf{73.9} \\
			\hline
		\end{tabular}
		\vspace{8pt}
		\caption{Quantitative results on the ScanNetV2~\cite{dai2017scannet} benchmark in terms of mAP, mAP@0.5 and mAP@0.25, respectively. Our approach outperforms previous methods by a significant margin. ScanNetV2 benchmark data is accessed on 11/14/2019.}
		\label{tab:performance_scannet}
	\end{table}

	\subsection{Quantitative Evaluation}
	Based on the public datasets, our methods are quantitatively compared with a number of representative existing methods, including SGPN~\cite{wang2018sgpn}, 3D-SIS~\cite{hou20193d}, PanopticFusion~\cite{narita2019panopticfusion}, 3D-BoNet~\cite{yang2019learning}, MTML~\cite{lahoud20193d}, ASIS~\cite{wang2019associatively} and JSIS3D~\cite{pham2019jsis3d}. 

	\begin{table}[tbp]
		\centering
		\begin{tabular}{|c|c|c|}
			\hline
			& mPrec  & mRec  \\
			\hline
			PartNet~\cite{mo2019partnet} & 56.4 & 43.4  \\
			\hline
			ASIS~\cite{wang2019associatively} & 63.6 & 47.5  \\
			\hline
			3D-BoNet~\cite{yang2019learning} & {65.6} & {47.6}  \\
			\hline
			OccuSeg & \textbf{72.8} & \textbf{60.3}  \\
			\hline
		\end{tabular}
		\vspace{8pt}
		\caption{Comparison on the S3DIS~\cite{armeni20163d} dataset. Our method outperforms previous methods in terms of mean Precision (mPrec) and mean recall (mRec) with an IoU threshold of $0.5$. }
		\label{tab:precision_recall_s3dis}
	\end{table}

	\begin{table*}[tbp]
		\centering
		\small
		\begin{tabular}{|c|c|cccccccccc|}
			\hline
			Method   & \textbf{mAP@0.5} &{wall}  & {floor} & cabinet &bed & {chair}& {sofa}& {table}  & {desk} & {tv} &{prop} \\
			\hline
			MT-PNet~\cite{pham2019jsis3d} & 8.5 & 13.1 & 27.3 & 0.0 &	15.0 &	21.2 &	0.0 &	0.7 & 0.0 &	6.0 &	{2.0}  
			\\
			\hline
			MLS-CRF~\cite{pham2019jsis3d} &12.1  & 13.9 & 44.5 & 0.0  & 32.9 & 12.9 & 0.0 & 5.7 &10.8  & 0.0 & 0.8   \\ \hline
			OccuSeg  &  \textbf{47.1} &	\textbf{39.0} &	\textbf{93.8} &	\textbf{5.7} & \textbf{66.7} &	\textbf{91.3} &	\textbf{8.7} &\textbf{50.0} &	\textbf{31.6} &	\textbf{76.9} &	\textbf{7.14} \\
			\hline
		\end{tabular}
		\vspace{8pt}
		\caption{Quantitative results on the SceneNN~\cite{hua2016scenenn} dataset in terms of mAP@0.5 score of each class. Our approach achieves the best performance for all the 10 classes.
		}
		\label{tab:perclass_iou_scenenn}
	\end{table*}

	\newcommand{\tabincell}[2]{\begin{tabular}{@{}#1@{}}#2\end{tabular}}  
	\begin{table}[tbp]
		\centering
		\small
		\begin{tabular}{|c|c|c|}
			\hline
			& Details & Total  \\
			\hline
			SGPN~\cite{wang2018sgpn} & \tabincell{c}{network(GPU): 650 \\
				group merging(CPU): 46562\\
				block merging(CPU): 2221}  &  49433 \\
			\hline
			ASIS~\cite{wang2019associatively} & \tabincell{c}{network(GPU): 650 \\
				mean shift(CPU): 53886\\
				block merging(CPU): 2221}  &  56757 \\
			\hline
			GSPN~\cite{yi2019gspn} & \tabincell{c}{network(GPU): 500\\
				point sampling(GPU): 2995\\
				neighbour search(CPU): 468}  &  3963 \\
			\hline
			3D-SIS~\cite{hou20193d} & \tabincell{c}{network (GPU+CPU):
				38841}  &  38841 \\
			\hline
			3D-BoNet~\cite{yang2019learning} & \tabincell{c}{network(GPU): 650\\
				SCN (GPU parallel): 208\\
				block merging(CPU): 2221}  &  2871 \\
			\hline
			OccuSeg & \tabincell{c}{network(GPU): 59 \\
				supervoxel(CPU): 375\\
				clustering(GPU+CPU): 160}  &  \textbf{594} \\
			
			\hline
		\end{tabular}
		\vspace{8pt}
		\caption{The processing time (seconds) on the validation set of ScanNetV2~\cite{dai2017scannet}. Note that all the other methods are evaluated based on their released codes according to~\cite{yang2019learning}.}
		\label{tab:efficiency}
	\end{table}
	\begin{table}[tbp]
		\centering
		\small
		\begin{tabular}{|c|c|c|c|}
			\hline
			& mAP  & mAP@0.5 & mAP@0.25\\
			\hline
			w/o\_feature & 36.7 & 51.8  & 62.6 \\
			\hline
			w/o\_spatial & 42.8 & 58.5 & 69.7\\
			\hline
			w/o\_occupancy& 40.9 & 55.7 & 67.4\\
			\hline
			OccuSeg & 44.2 & 60.7 & 71.9\\
			\hline
		\end{tabular}
		\vspace{8pt}
		\caption{Ablation study of each component of our method on the ScanNetV2 validation split, in terms of mAP, mAP@0.5 and mAP@0.25.}
		\label{tab:ablation}
	\end{table}
	\myparagraph{ScanNetV2.} We follow the benchmark~\cite{dai2017scannet} to use the mean average precision at overlap 0.25 (mAP@0.25), overlap 0.5 (mAP@0.5) and overlaps in the range $[0.5:0.95:0.05]$ (mAP) as evaluation metrics. Tab.~\ref{tab:perclass_iou_scannet} and Tab.~\ref{tab:performance_scannet} summarize the 
	per-class mAP and the overall performance, respectively. 
	Overall, our method achieves a significant margin on all the three metrics, especially the hardest mAP metric, indicating the effectiveness of our method for 3D instance segmentation.

	\myparagraph{S3DIS.} Following the previous methods~\cite{yang2019learning,wang2019associatively}, we employ the 6-fold cross validation and use the mean precision (mPrec) / mean recall (mRec) with an IoU threshold 0.5 to evaluate our method in the S3DIS dataset. 
	As shown in Tab.~\ref{tab:precision_recall_s3dis}, our scheme outperforms all the previous methods by a significant margin in terms of both mPrec and mRec, indicating its ability to segment more instances precisely.

	\myparagraph{SceneNN.} Similar to previous work~\cite{pham2019jsis3d}, mAP@0.5 metric is adopted to evaluate our approach in the SceneNN dataset. 
	As presented in Tab.~\ref{tab:perclass_iou_scenenn}, even using only 50 scans for training, our approach outperforms the previous method~\cite{pham2019jsis3d} by a significant margin (35 for mAP@0.5), which illustrates the effectiveness of our approach under small datasets.

	\subsection{Complexity Analysis}
	Maintaining high efficiency plays a vital role when applying 3D instance segmentation to mixed reality or robotics applications. Similar to the evaluation in \cite{yang2019learning}, we made a comparison of the processing times on the 312 scans of indoor environments from the validation split of ScanNetV2. 
	Both the proposal-based methods (3DSIS~\cite{hou20193d}, GSPN~\cite{yi2019gspn} and 3D-BoNet~\cite{yang2019learning}) and the proposal-free methods (SGPN~\cite{wang2018sgpn} ASIS~\cite{wang2019associatively}) are concerned.
	The processing time of the full pipeline and the main stages are respectively reported in Tab.~\ref{tab:efficiency}.
	Remarkably, our method is more than $\mathbf{4\times}$ faster than the existed most efficient approach 3D-BoNet~\cite{yang2019learning}.

	\subsection{Ablation Study}
	Here, we evaluate the individual components of our method on the ScanNetV2 validation split. 
	Let \textit{w/o\_feature} and \textit{w/o\_spatial} denote the variations of our method without the feature embedding or spatial feature embedding, respectively. 
	%
	%
	To evaluate the influence of the novel occupancy signal, we disable the occupancy prediction in the learning stage and set the occupancy ratio $r = 1$ for all vertices in Eqn.~\ref{eqn_weight} during the clustering stage, denoted as \textit{w/o\_occupancy}.
	
	The quantitative comparison results of all the variations of our method are provided in Tab.~\ref{tab:ablation}, which demonstrate that the proposed occupancy aware scheme helps to improve the overall quality of 3D instance segmentation.


	\section{Discussion and Conclusion}
	We presented OccuSeg, an occupancy-aware instance segmentation method for 3D scenes.
	Our learning stage leverages feature embedding and spatial embedding, as well as a novel 3D occupancy signal to imply the inherent property of 3D objects.
	The occupancy signal further guides our graph-based clustering stage to correctly merge hard samples and prohibit over-segmented clusters.
	Extensive experimental results demonstrate the effectiveness of our method, which outperforms previous methods by a significant margin and retains high efficiency.
	%
	\zt{In the future work, we will improve our method by incorporating tailored designs for partially reconstructed objects.}
	Also, we intend to investigate the sub-object level 3D instance segmentation and further improve the efficiency, enabling the practical usage of high quality 3D instance segmentation for tremendous applications in AR/VR, gaming and mobile robots. 
	
	

	{\small
		\bibliographystyle{ieee_fullname}
		\bibliography{ms}

\begin{thebibliography}{10}\itemsep=-1pt

\bibitem{armeni20163d}
Iro Armeni, Ozan Sener, Amir~R Zamir, Helen Jiang, Ioannis Brilakis, Martin
  Fischer, and Silvio Savarese.
\newblock 3d semantic parsing of large-scale indoor spaces.
\newblock In {\em Proceedings of the IEEE Conference on Computer Vision and
  Pattern Recognition}, pages 1534--1543, 2016.

\bibitem{choy20194d}
Christopher Choy, JunYoung Gwak, and Silvio Savarese.
\newblock 4d spatio-temporal convnets: Minkowski convolutional neural networks.
\newblock {\em arXiv preprint arXiv:1904.08755}, 2019.

\bibitem{christoph2014object}
Simon Christoph~Stein, Markus Schoeler, Jeremie Papon, and Florentin Worgotter.
\newblock Object partitioning using local convexity.
\newblock In {\em Proceedings of the IEEE Conference on Computer Vision and
  Pattern Recognition}, pages 304--311, 2014.

\bibitem{dai2017scannet}
Angela Dai, Angel~X Chang, Manolis Savva, Maciej Halber, Thomas Funkhouser, and
  Matthias Nie{\ss}ner.
\newblock Scannet: Richly-annotated 3d reconstructions of indoor scenes.
\newblock In {\em Proceedings of the IEEE Conference on Computer Vision and
  Pattern Recognition}, pages 5828--5839, 2017.

\bibitem{dai2017bundlefusion}
Angela Dai, Matthias Nie{\ss}ner, Michael Zollh{\"o}fer, Shahram Izadi, and
  Christian Theobalt.
\newblock Bundlefusion: Real-time globally consistent 3d reconstruction using
  on-the-fly surface reintegration.
\newblock {\em ACM Transactions on Graphics (TOG)}, 36(3):24, 2017.

\bibitem{daiECCV2016_IFCN}
Jifeng Dai, Kaiming He, Yi Li, Shaoqing Ren, and Jian Sun.
\newblock Instance-sensitive fully convolutional networks.
\newblock In Bastian Leibe, Jiri Matas, Nicu Sebe, and Max Welling, editors,
  {\em Computer Vision -- ECCV 2016}, pages 534--549, Cham, 2016. Springer
  International Publishing.

\bibitem{de2017semantic}
Bert De~Brabandere, Davy Neven, and Luc Van~Gool.
\newblock Semantic instance segmentation with a discriminative loss function.
\newblock {\em arXiv preprint arXiv:1708.02551}, 2017.

\bibitem{vote3deep}
M. {Engelcke}, D. {Rao}, D.~Z. {Wang}, C.~H. {Tong}, and I. {Posner}.
\newblock Vote3deep: Fast object detection in 3d point clouds using efficient
  convolutional neural networks.
\newblock In {\em 2017 IEEE International Conference on Robotics and Automation
  (ICRA)}, pages 1355--1361, 2017.

\bibitem{Fathi17_dml}
Alireza Fathi, Zbigniew Wojna, Vivek Rathod, Peng Wang, Hyun~Oh Song, Sergio
  Guadarrama, and Kevin~P. Murphy.
\newblock Semantic instance segmentation via deep metric learning.
\newblock {\em CoRR}, abs/1703.10277, 2017.

\bibitem{felzenszwalb2004efficient}
Pedro~F Felzenszwalb and Daniel~P Huttenlocher.
\newblock Efficient graph-based image segmentation.
\newblock {\em International journal of computer vision}, 59(2):167--181, 2004.

\bibitem{fastRCNN}
Ross Girshick.
\newblock Fast r-cnn.
\newblock In {\em The IEEE International Conference on Computer Vision (ICCV)},
  December 2015.

\bibitem{goodfellow2016deep}
Ian Goodfellow, Yoshua Bengio, and Aaron Courville.
\newblock {\em Deep learning}.
\newblock MIT press, 2016.

\bibitem{graham2014spatially}
Benjamin Graham.
\newblock Spatially-sparse convolutional neural networks.
\newblock {\em arXiv preprint arXiv:1409.6070}, 2014.

\bibitem{graham20183d}
Benjamin Graham, Martin Engelcke, and Laurens van~der Maaten.
\newblock 3d semantic segmentation with submanifold sparse convolutional
  networks.
\newblock In {\em Proceedings of the IEEE Conference on Computer Vision and
  Pattern Recognition}, pages 9224--9232, 2018.

\bibitem{han2018flashfusion}
Lei Han and Lu Fang.
\newblock Flashfusion: Real-time globally consistent dense 3d reconstruction
  using cpu computing.
\newblock In {\em Robotics: Science and Systems}, 2018.

\bibitem{BoundaryIS_2017}
Zeeshan Hayder, Xuming He, and Mathieu Salzmann.
\newblock Boundary-aware instance segmentation.
\newblock In {\em The IEEE Conference on Computer Vision and Pattern
  Recognition (CVPR)}, July 2017.

\bibitem{he2017mask}
Kaiming He, Georgia Gkioxari, Piotr Doll{\'a}r, and Ross Girshick.
\newblock Mask r-cnn.
\newblock In {\em Proceedings of the IEEE international conference on computer
  vision}, pages 2961--2969, 2017.

\bibitem{hou20193d}
Ji Hou, Angela Dai, and Matthias Nie{\ss}ner.
\newblock 3d-sis: 3d semantic instance segmentation of rgb-d scans.
\newblock In {\em Proceedings of the IEEE Conference on Computer Vision and
  Pattern Recognition}, pages 4421--4430, 2019.

\bibitem{hua2016scenenn}
Binh-Son Hua, Quang-Hieu Pham, Duc~Thanh Nguyen, Minh-Khoi Tran, Lap-Fai Yu,
  and Sai-Kit Yeung.
\newblock Scenenn: A scene meshes dataset with annotations.
\newblock In {\em 2016 Fourth International Conference on 3D Vision (3DV)},
  pages 92--101. IEEE, 2016.

\bibitem{han2019Semantic}
Anonymous~(Attached in~the~supplementary material).
\newblock Realtime semantic 3d perception for immersive augmented reality.
\newblock In {\em conditionally accepted by IEEE VR/TVCG}. IEEE, 2019.

\bibitem{Kong_2018_CVPR}
Shu Kong and Charless~C. Fowlkes.
\newblock Recurrent pixel embedding for instance grouping.
\newblock In {\em The IEEE Conference on Computer Vision and Pattern
  Recognition (CVPR)}, June 2018.

\bibitem{lahoud20193d}
Jean Lahoud, Bernard Ghanem, Marc Pollefeys, and Martin~R Oswald.
\newblock 3d instance segmentation via multi-task metric learning.
\newblock {\em arXiv preprint arXiv:1906.08650}, 2019.

\bibitem{superpoint}
L. {Landrieu} and M. {Simonovsky}.
\newblock Large-scale point cloud semantic segmentation with superpoint graphs.
\newblock In {\em 2018 IEEE/CVF Conference on Computer Vision and Pattern
  Recognition}, pages 4558--4567, 2018.

\bibitem{leibe2008robust}
Bastian Leibe, Ale{\v{s}} Leonardis, and Bernt Schiele.
\newblock Robust object detection with interleaved categorization and
  segmentation.
\newblock {\em International journal of computer vision}, 77(1-3):259--289,
  2008.

\bibitem{LiFCN_CVPR2017}
Yi Li, Haozhi Qi, Jifeng Dai, Xiangyang Ji, and Yichen Wei.
\newblock Fully convolutional instance-aware semantic segmentation.
\newblock In {\em The IEEE Conference on Computer Vision and Pattern
  Recognition (CVPR)}, July 2017.

\bibitem{liang2017proposal}
Xiaodan Liang, Liang Lin, Yunchao Wei, Xiaohui Shen, Jianchao Yang, and
  Shuicheng Yan.
\newblock Proposal-free network for instance-level object segmentation.
\newblock {\em IEEE transactions on pattern analysis and machine intelligence},
  40(12):2978--2991, 2017.

\bibitem{liang20193d}
Zhidong Liang, Ming Yang, and Chunxiang Wang.
\newblock 3d graph embedding learning with a structure-aware loss function for
  point cloud semantic instance segmentation.
\newblock {\em arXiv preprint arXiv:1902.05247}, 2019.

\bibitem{liu2019masc}
Chen Liu and Yasutaka Furukawa.
\newblock Masc: Multi-scale affinity with sparse convolution for 3d instance
  segmentation.
\newblock {\em arXiv preprint arXiv:1902.04478}, 2019.

\bibitem{liu2018affinity}
Yiding Liu, Siyu Yang, Bin Li, Wengang Zhou, Jizheng Xu, Houqiang Li, and Yan
  Lu.
\newblock Affinity derivation and graph merge for instance segmentation.
\newblock In {\em Proceedings of the European Conference on Computer Vision
  (ECCV)}, pages 686--703, 2018.

\bibitem{mo2019partnet}
Kaichun Mo, Shilin Zhu, Angel~X Chang, Li Yi, Subarna Tripathi, Leonidas~J
  Guibas, and Hao Su.
\newblock Partnet: A large-scale benchmark for fine-grained and hierarchical
  part-level 3d object understanding.
\newblock In {\em Proceedings of the IEEE Conference on Computer Vision and
  Pattern Recognition}, pages 909--918, 2019.

\bibitem{narita2019panopticfusion}
Gaku Narita, Takashi Seno, Tomoya Ishikawa, and Yohsuke Kaji.
\newblock Panopticfusion: Online volumetric semantic mapping at the level of
  stuff and things.
\newblock {\em arXiv preprint arXiv:1903.01177}, 2019.

\bibitem{neven2019instance}
Davy Neven, Bert~De Brabandere, Marc Proesmans, and Luc~Van Gool.
\newblock Instance segmentation by jointly optimizing spatial embeddings and
  clustering bandwidth.
\newblock In {\em Proceedings of the IEEE Conference on Computer Vision and
  Pattern Recognition}, pages 8837--8845, 2019.

\bibitem{newcombe2011kinectfusion}
R.~A. {Newcombe}, S. {Izadi}, O. {Hilliges}, D. {Molyneaux}, D. {Kim}, A.~J.
  {Davison}, P. {Kohi}, J. {Shotton}, S. {Hodges}, and A. {Fitzgibbon}.
\newblock Kinectfusion: Real-time dense surface mapping and tracking.
\newblock In {\em 2011 10th IEEE International Symposium on Mixed and Augmented
  Reality}, pages 127--136, 2011.

\bibitem{niessner2013real}
Matthias Nießner, Michael Zollhöfer, Shahram Izadi, and Stamminger.
\newblock Real-time 3d reconstruction at scale using voxel hashing.
\newblock {\em ACM Transactions on Graphics (TOG)}, 32(6):169, 2013.

\bibitem{novotny2018semi}
David Novotny, Samuel Albanie, Diane Larlus, and Andrea Vedaldi.
\newblock Semi-convolutional operators for instance segmentation.
\newblock In {\em Proceedings of the European Conference on Computer Vision
  (ECCV)}, pages 86--102, 2018.

\bibitem{pham2019jsis3d}
Quang-Hieu Pham, Thanh Nguyen, Binh-Son Hua, Gemma Roig, and Sai-Kit Yeung.
\newblock Jsis3d: Joint semantic-instance segmentation of 3d point clouds with
  multi-task pointwise networks and multi-value conditional random fields.
\newblock In {\em Proceedings of the IEEE Conference on Computer Vision and
  Pattern Recognition}, pages 8827--8836, 2019.

\bibitem{qi2019deep}
Charles~R Qi, Or Litany, Kaiming He, and Leonidas~J Guibas.
\newblock Deep hough voting for 3d object detection in point clouds.
\newblock {\em arXiv preprint arXiv:1904.09664}, 2019.

\bibitem{qi2017pointnet}
Charles~R Qi, Hao Su, Kaichun Mo, and Leonidas~J Guibas.
\newblock Pointnet: Deep learning on point sets for 3d classification and
  segmentation.
\newblock In {\em Proceedings of the IEEE Conference on Computer Vision and
  Pattern Recognition}, pages 652--660, 2017.

\bibitem{qi2017pointnet2}
Charles~Ruizhongtai Qi, Li Yi, Hao Su, and Leonidas~J Guibas.
\newblock Pointnet++: Deep hierarchical feature learning on point sets in a
  metric space.
\newblock In {\em Advances in neural information processing systems}, pages
  5099--5108, 2017.

\bibitem{ronneberger2015u}
Olaf Ronneberger, Philipp Fischer, and Thomas Brox.
\newblock U-net: Convolutional networks for biomedical image segmentation.
\newblock In {\em International Conference on Medical image computing and
  computer-assisted intervention}, pages 234--241. Springer, 2015.

\bibitem{schuurmans2018efficient}
Mathijs Schuurmans, Maxim Berman, and Matthew~B Blaschko.
\newblock Efficient semantic image segmentation with superpixel pooling.
\newblock {\em arXiv preprint arXiv:1806.02705}, 2018.

\bibitem{tchapmi2017segcloud}
Lyne Tchapmi, Christopher Choy, Iro Armeni, JunYoung Gwak, and Silvio Savarese.
\newblock Segcloud: Semantic segmentation of 3d point clouds.
\newblock In {\em 2017 International Conference on 3D Vision (3DV)}, pages
  537--547. IEEE, 2017.

\bibitem{wang2018sgpn}
Weiyue Wang, Ronald Yu, Qiangui Huang, and Ulrich Neumann.
\newblock Sgpn: Similarity group proposal network for 3d point cloud instance
  segmentation.
\newblock In {\em Proceedings of the IEEE Conference on Computer Vision and
  Pattern Recognition}, pages 2569--2578, 2018.

\bibitem{wang2019associatively}
Xinlong Wang, Shu Liu, Xiaoyong Shen, Chunhua Shen, and Jiaya Jia.
\newblock Associatively segmenting instances and semantics in point clouds.
\newblock In {\em Proceedings of the IEEE Conference on Computer Vision and
  Pattern Recognition}, pages 4096--4105, 2019.

\bibitem{wu20153d}
Zhirong Wu, Shuran Song, Aditya Khosla, Fisher Yu, Linguang Zhang, Xiaoou Tang,
  and Jianxiong Xiao.
\newblock 3d shapenets: A deep representation for volumetric shapes.
\newblock In {\em Proceedings of the IEEE conference on computer vision and
  pattern recognition}, pages 1912--1920, 2015.

\bibitem{xiong2019upsnet}
Yuwen Xiong, Renjie Liao, Hengshuang Zhao, Rui Hu, Min Bai, Ersin Yumer, and
  Raquel Urtasun.
\newblock Upsnet: A unified panoptic segmentation network.
\newblock In {\em Proceedings of the IEEE Conference on Computer Vision and
  Pattern Recognition}, pages 8818--8826, 2019.

\bibitem{LanXu2019}
L. {Xu}, Z. {Su}, L. {Han}, T. {Yu}, Y. {Liu}, and L. {FANG}.
\newblock Unstructuredfusion: Realtime 4d geometry and texture reconstruction
  using commercialrgbd cameras.
\newblock {\em IEEE Transactions on Pattern Analysis and Machine Intelligence},
  pages 1--1, 2019.

\bibitem{yang2019learning}
Bo Yang, Jianan Wang, Ronald Clark, Qingyong Hu, Sen Wang, Andrew Markham, and
  Niki Trigoni.
\newblock Learning object bounding boxes for 3d instance segmentation on point
  clouds.
\newblock {\em arXiv preprint arXiv:1906.01140}, 2019.

\bibitem{yang2017semantic}
Shichao Yang, Yulan Huang, and Sebastian Scherer.
\newblock Semantic 3d occupancy mapping through efficient high order crfs.
\newblock In {\em 2017 IEEE/RSJ International Conference on Intelligent Robots
  and Systems (IROS)}, pages 590--597. IEEE, 2017.

\bibitem{yi2019gspn}
Li Yi, Wang Zhao, He Wang, Minhyuk Sung, and Leonidas~J Guibas.
\newblock Gspn: Generative shape proposal network for 3d instance segmentation
  in point cloud.
\newblock In {\em Proceedings of the IEEE Conference on Computer Vision and
  Pattern Recognition}, pages 3947--3956, 2019.

\end{thebibliography}
	}
	
	\newpage
	\appendix
		\twocolumn[{%
		\renewcommand\twocolumn[1][]{#1}%
		\maketitle
		\thispagestyle{empty}
		\begin{center}
			\centering
			\def\svgwidth{\textwidth}
	\executeiffilenewer{inkscape/network3.svg}{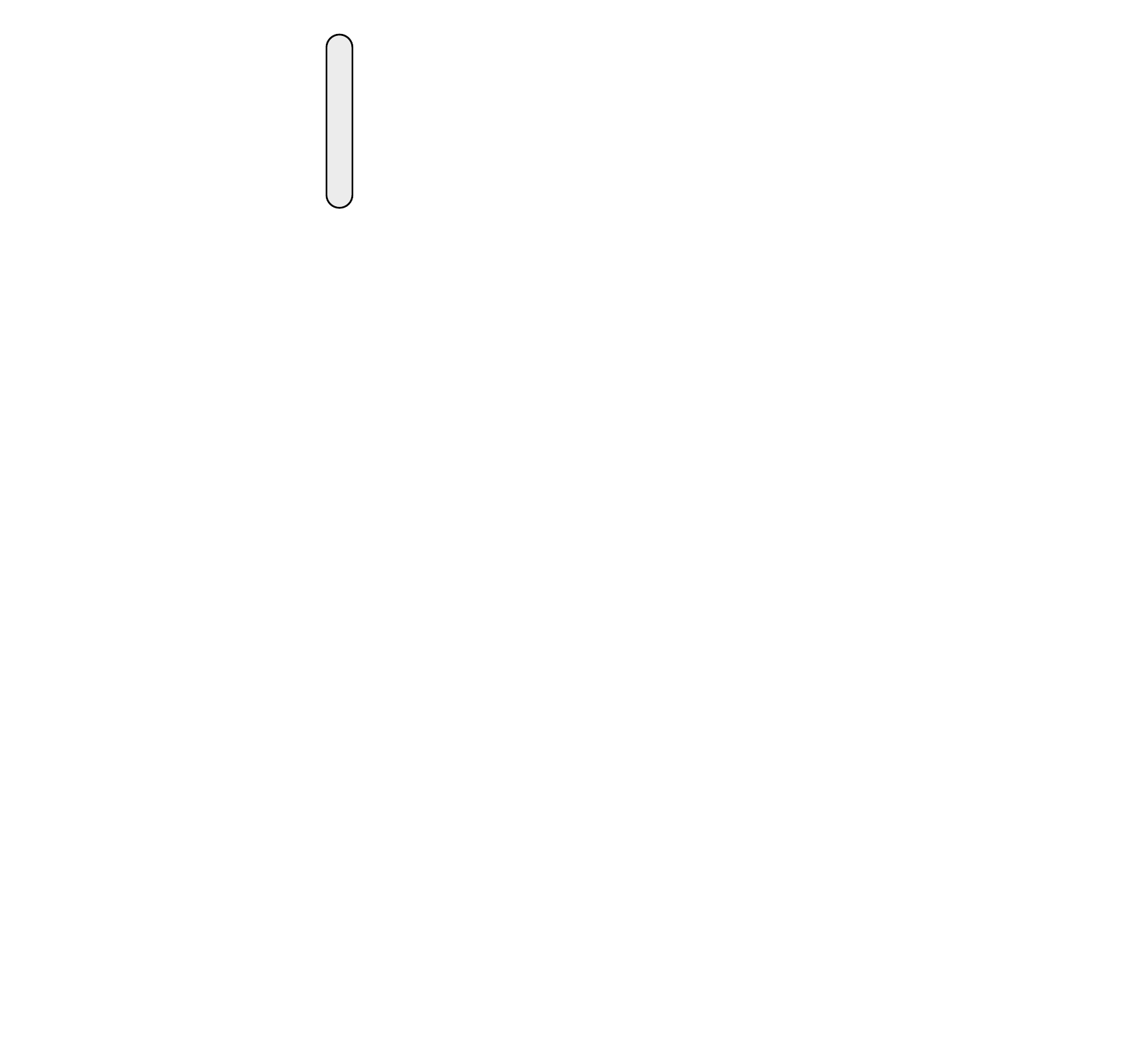}%
	{inkscape -z -D --file=inkscape/network3.svg %
		--export-pdf=inkscape/network3.pdf --export-latex}%
\begingroup%
  \makeatletter%
  \providecommand\color[2][]{%
    \errmessage{(Inkscape) Color is used for the text in Inkscape, but the package 'color.sty' is not loaded}%
    \renewcommand\color[2][]{}%
  }%
  \providecommand\transparent[1]{%
    \errmessage{(Inkscape) Transparency is used (non-zero) for the text in Inkscape, but the package 'transparent.sty' is not loaded}%
    \renewcommand\transparent[1]{}%
  }%
  \providecommand\rotatebox[2]{#2}%
  \newcommand*\fsize{\dimexpr\f@size pt\relax}%
  \newcommand*\lineheight[1]{\fontsize{\fsize}{#1\fsize}\selectfont}%
  \ifx\svgwidth\undefined%
    \setlength{\unitlength}{496.88503943bp}%
    \ifx\svgscale\undefined%
      \relax%
    \else%
      \setlength{\unitlength}{\unitlength * \real{\svgscale}}%
    \fi%
  \else%
    \setlength{\unitlength}{\svgwidth}%
  \fi%
  \global\let\svgwidth\undefined%
  \global\let\svgscale\undefined%
  \makeatother%
  \begin{picture}(1,0.90896566)%
    \lineheight{1}%
    \setlength\tabcolsep{0pt}%
    \put(0,0){\includegraphics[width=\unitlength,page=1]{network3.pdf}}%
    \put(0.30247414,0.80339009){\color[rgb]{0,0,0}\rotatebox{90}{\makebox(0,0)[t]{\lineheight{1.25}\smash{\begin{tabular}[t]{c}Input ($N*3$)\end{tabular}}}}}%
    \put(0,0){\includegraphics[width=\unitlength,page=2]{network3.pdf}}%
    \put(0.34775625,0.8033901){\color[rgb]{0,0,0}\rotatebox{90}{\makebox(0,0)[t]{\lineheight{1.25}\smash{\begin{tabular}[t]{c}UNet Module\end{tabular}}}}}%
    \put(0,0){\includegraphics[width=\unitlength,page=3]{network3.pdf}}%
    \put(0.39142423,0.79992417){\color[rgb]{0,0,0}\rotatebox{90}{\makebox(0,0)[t]{\lineheight{1.25}\smash{\begin{tabular}[t]{c}Linear Layers\end{tabular}}}}}%
    \put(0,0){\includegraphics[width=\unitlength,page=4]{network3.pdf}}%
    \put(0.57903186,0.83058827){\color[rgb]{0,0,0}\makebox(0,0)[t]{\lineheight{1.25}\smash{\begin{tabular}[t]{c}Semantic Term ($N*K$)\end{tabular}}}}%
    \put(0,0){\includegraphics[width=\unitlength,page=5]{network3.pdf}}%
    \put(0.57902263,0.86548602){\color[rgb]{0,0,0}\makebox(0,0)[t]{\lineheight{1.25}\smash{\begin{tabular}[t]{c}Spatial Term ($N*3$)\end{tabular}}}}%
    \put(0,0){\includegraphics[width=\unitlength,page=6]{network3.pdf}}%
    \put(0.57912117,0.79624945){\color[rgb]{0,0,0}\makebox(0,0)[t]{\lineheight{1.25}\smash{\begin{tabular}[t]{c}Covariance ($N*2$)\end{tabular}}}}%
    \put(0,0){\includegraphics[width=\unitlength,page=7]{network3.pdf}}%
    \put(0.57860236,0.75039402){\color[rgb]{0,0,0}\makebox(0,0)[t]{\lineheight{1.25}\smash{\begin{tabular}[t]{c}Occupancy ($N*1$)\end{tabular}}}}%
    \put(0,0){\includegraphics[width=\unitlength,page=8]{network3.pdf}}%
    \put(0.57860236,0.71605517){\color[rgb]{0,0,0}\makebox(0,0)[t]{\lineheight{1.25}\smash{\begin{tabular}[t]{c}Semantic Seg. ($N*C$)\end{tabular}}}}%
    \put(0,0){\includegraphics[width=\unitlength,page=9]{network3.pdf}}%
    \put(0.57999397,0.88958893){\color[rgb]{0,0,0}\makebox(0,0)[t]{\lineheight{1.25}\smash{\begin{tabular}[t]{c}\textbf{Learned Embeddings}\end{tabular}}}}%
    \put(0,0){\includegraphics[width=\unitlength,page=10]{network3.pdf}}%
    \put(0.07319258,0.6182608){\color[rgb]{0,0,0}\rotatebox{90}{\makebox(0,0)[t]{\lineheight{1.25}\smash{\begin{tabular}[t]{c}{\tiny SSC, K3S1, 32}\end{tabular}}}}}%
    \put(0,0){\includegraphics[width=\unitlength,page=11]{network3.pdf}}%
    \put(0.09583363,0.6182608){\color[rgb]{0,0,0}\rotatebox{90}{\makebox(0,0)[t]{\lineheight{1.25}\smash{\begin{tabular}[t]{c}{\tiny SSC, K3S1, 32}\end{tabular}}}}}%
    \put(0,0){\includegraphics[width=\unitlength,page=12]{network3.pdf}}%
    \put(0.11847469,0.6182608){\color[rgb]{0,0,0}\rotatebox{90}{\makebox(0,0)[t]{\lineheight{1.25}\smash{\begin{tabular}[t]{c}{\tiny SSC, K3S1, 32}\end{tabular}}}}}%
    \put(0,0){\includegraphics[width=\unitlength,page=13]{network3.pdf}}%
    \put(0.14111574,0.52559617){\color[rgb]{0,0,0}\rotatebox{90}{\makebox(0,0)[t]{\lineheight{1.25}\smash{\begin{tabular}[t]{c}{\tiny SC, K3S2, 64}\end{tabular}}}}}%
    \put(0,0){\includegraphics[width=\unitlength,page=14]{network3.pdf}}%
    \put(0.16375679,0.52559617){\color[rgb]{0,0,0}\rotatebox{90}{\makebox(0,0)[t]{\lineheight{1.25}\smash{\begin{tabular}[t]{c}{\tiny SSC, K3S1, 64}\end{tabular}}}}}%
    \put(0,0){\includegraphics[width=\unitlength,page=15]{network3.pdf}}%
    \put(0.18639784,0.52559617){\color[rgb]{0,0,0}\rotatebox{90}{\makebox(0,0)[t]{\lineheight{1.25}\smash{\begin{tabular}[t]{c}{\tiny SSC, K3S1, 64}\end{tabular}}}}}%
    \put(0,0){\includegraphics[width=\unitlength,page=16]{network3.pdf}}%
    \put(0.2090389,0.43293154){\color[rgb]{0,0,0}\rotatebox{90}{\makebox(0,0)[t]{\lineheight{1.25}\smash{\begin{tabular}[t]{c}{\tiny SC, K3S2, 96}\end{tabular}}}}}%
    \put(0,0){\includegraphics[width=\unitlength,page=17]{network3.pdf}}%
    \put(0.23167995,0.43293154){\color[rgb]{0,0,0}\rotatebox{90}{\makebox(0,0)[t]{\lineheight{1.25}\smash{\begin{tabular}[t]{c}{\tiny SSC, K3S1, 96}\end{tabular}}}}}%
    \put(0,0){\includegraphics[width=\unitlength,page=18]{network3.pdf}}%
    \put(0.254321,0.43293154){\color[rgb]{0,0,0}\rotatebox{90}{\makebox(0,0)[t]{\lineheight{1.25}\smash{\begin{tabular}[t]{c}{\tiny SSC, K3S1, 96}\end{tabular}}}}}%
    \put(0,0){\includegraphics[width=\unitlength,page=19]{network3.pdf}}%
    \put(0.27696206,0.3402669){\color[rgb]{0,0,0}\rotatebox{90}{\makebox(0,0)[t]{\lineheight{1.25}\smash{\begin{tabular}[t]{c}{\tiny SC, K3S2, 128}\end{tabular}}}}}%
    \put(0,0){\includegraphics[width=\unitlength,page=20]{network3.pdf}}%
    \put(0.2996031,0.34026687){\color[rgb]{0,0,0}\rotatebox{90}{\makebox(0,0)[t]{\lineheight{1.25}\smash{\begin{tabular}[t]{c}{\tiny SSC, K3S1, 128}\end{tabular}}}}}%
    \put(0,0){\includegraphics[width=\unitlength,page=21]{network3.pdf}}%
    \put(0.32224415,0.34026687){\color[rgb]{0,0,0}\rotatebox{90}{\makebox(0,0)[t]{\lineheight{1.25}\smash{\begin{tabular}[t]{c}{\tiny SSC, K3S1, 128}\end{tabular}}}}}%
    \put(0,0){\includegraphics[width=\unitlength,page=22]{network3.pdf}}%
    \put(0.34488521,0.24760228){\color[rgb]{0,0,0}\rotatebox{90}{\makebox(0,0)[t]{\lineheight{1.25}\smash{\begin{tabular}[t]{c}{\tiny SC, K3S2, 160}\end{tabular}}}}}%
    \put(0,0){\includegraphics[width=\unitlength,page=23]{network3.pdf}}%
    \put(0.36752626,0.24760225){\color[rgb]{0,0,0}\rotatebox{90}{\makebox(0,0)[t]{\lineheight{1.25}\smash{\begin{tabular}[t]{c}{\tiny SSC, K3S1, 160}\end{tabular}}}}}%
    \put(0,0){\includegraphics[width=\unitlength,page=24]{network3.pdf}}%
    \put(0.39016731,0.24760225){\color[rgb]{0,0,0}\rotatebox{90}{\makebox(0,0)[t]{\lineheight{1.25}\smash{\begin{tabular}[t]{c}{\tiny SSC, K3S1, 160}\end{tabular}}}}}%
    \put(0,0){\includegraphics[width=\unitlength,page=25]{network3.pdf}}%
    \put(0.41280835,0.15493761){\color[rgb]{0,0,0}\rotatebox{90}{\makebox(0,0)[t]{\lineheight{1.25}\smash{\begin{tabular}[t]{c}{\tiny SC, K3S2, 192}\end{tabular}}}}}%
    \put(0,0){\includegraphics[width=\unitlength,page=26]{network3.pdf}}%
    \put(0.4354494,0.15493764){\color[rgb]{0,0,0}\rotatebox{90}{\makebox(0,0)[t]{\lineheight{1.25}\smash{\begin{tabular}[t]{c}{\tiny SSC, K3S1, 192}\end{tabular}}}}}%
    \put(0,0){\includegraphics[width=\unitlength,page=27]{network3.pdf}}%
    \put(0.45809047,0.15493764){\color[rgb]{0,0,0}\rotatebox{90}{\makebox(0,0)[t]{\lineheight{1.25}\smash{\begin{tabular}[t]{c}{\tiny SSC, K3S1, 192}\end{tabular}}}}}%
    \put(0,0){\includegraphics[width=\unitlength,page=28]{network3.pdf}}%
    \put(0.48073151,0.06227299){\color[rgb]{0,0,0}\rotatebox{90}{\makebox(0,0)[t]{\lineheight{1.25}\smash{\begin{tabular}[t]{c}{\tiny SC, K3S2, 224}\end{tabular}}}}}%
    \put(0,0){\includegraphics[width=\unitlength,page=29]{network3.pdf}}%
    \put(0.50337255,0.06227297){\color[rgb]{0,0,0}\rotatebox{90}{\makebox(0,0)[t]{\lineheight{1.25}\smash{\begin{tabular}[t]{c}{\tiny SSC, K3S1, 224}\end{tabular}}}}}%
    \put(0,0){\includegraphics[width=\unitlength,page=30]{network3.pdf}}%
    \put(0.52601362,0.06227297){\color[rgb]{0,0,0}\rotatebox{90}{\makebox(0,0)[t]{\lineheight{1.25}\smash{\begin{tabular}[t]{c}{\tiny SSC, K3S1, 224}\end{tabular}}}}}%
    \put(0,0){\includegraphics[width=\unitlength,page=31]{network3.pdf}}%
    \put(0.54865467,0.06227297){\color[rgb]{0,0,0}\rotatebox{90}{\makebox(0,0)[t]{\lineheight{1.25}\smash{\begin{tabular}[t]{c}{\tiny Deconv, K3S2, 224}\end{tabular}}}}}%
    \put(0,0){\includegraphics[width=\unitlength,page=32]{network3.pdf}}%
    \put(0.57129571,0.15493764){\color[rgb]{0,0,0}\rotatebox{90}{\makebox(0,0)[t]{\lineheight{1.25}\smash{\begin{tabular}[t]{c}{\tiny SSC, K3S1, 192}\end{tabular}}}}}%
    \put(0,0){\includegraphics[width=\unitlength,page=33]{network3.pdf}}%
    \put(0.59393676,0.15493764){\color[rgb]{0,0,0}\rotatebox{90}{\makebox(0,0)[t]{\lineheight{1.25}\smash{\begin{tabular}[t]{c}{\tiny SSC, K3S1, 192}\end{tabular}}}}}%
    \put(0,0){\includegraphics[width=\unitlength,page=34]{network3.pdf}}%
    \put(0.61657783,0.15493764){\color[rgb]{0,0,0}\rotatebox{90}{\makebox(0,0)[t]{\lineheight{1.25}\smash{\begin{tabular}[t]{c}{\tiny Deconv, K3S2, 192}\end{tabular}}}}}%
    \put(0,0){\includegraphics[width=\unitlength,page=35]{network3.pdf}}%
    \put(0.63921889,0.24760225){\color[rgb]{0,0,0}\rotatebox{90}{\makebox(0,0)[t]{\lineheight{1.25}\smash{\begin{tabular}[t]{c}{\tiny SSC, K3S1, 160}\end{tabular}}}}}%
    \put(0,0){\includegraphics[width=\unitlength,page=36]{network3.pdf}}%
    \put(0.66186,0.24760225){\color[rgb]{0,0,0}\rotatebox{90}{\makebox(0,0)[t]{\lineheight{1.25}\smash{\begin{tabular}[t]{c}{\tiny SSC, K3S1, 160}\end{tabular}}}}}%
    \put(0,0){\includegraphics[width=\unitlength,page=37]{network3.pdf}}%
    \put(0.68450105,0.24760225){\color[rgb]{0,0,0}\rotatebox{90}{\makebox(0,0)[t]{\lineheight{1.25}\smash{\begin{tabular}[t]{c}{\tiny Deconv, K3S2, 160}\end{tabular}}}}}%
    \put(0,0){\includegraphics[width=\unitlength,page=38]{network3.pdf}}%
    \put(0.70714204,0.34026687){\color[rgb]{0,0,0}\rotatebox{90}{\makebox(0,0)[t]{\lineheight{1.25}\smash{\begin{tabular}[t]{c}{\tiny SSC, K3S1, 128}\end{tabular}}}}}%
    \put(0,0){\includegraphics[width=\unitlength,page=39]{network3.pdf}}%
    \put(0.7297831,0.34026687){\color[rgb]{0,0,0}\rotatebox{90}{\makebox(0,0)[t]{\lineheight{1.25}\smash{\begin{tabular}[t]{c}{\tiny SSC, K3S1, 128}\end{tabular}}}}}%
    \put(0,0){\includegraphics[width=\unitlength,page=40]{network3.pdf}}%
    \put(0.7524242,0.34026687){\color[rgb]{0,0,0}\rotatebox{90}{\makebox(0,0)[t]{\lineheight{1.25}\smash{\begin{tabular}[t]{c}{\tiny Deconv, K3S2, 128}\end{tabular}}}}}%
    \put(0,0){\includegraphics[width=\unitlength,page=41]{network3.pdf}}%
    \put(0.77506526,0.43293154){\color[rgb]{0,0,0}\rotatebox{90}{\makebox(0,0)[t]{\lineheight{1.25}\smash{\begin{tabular}[t]{c}{\tiny SSC, K3S1, 96}\end{tabular}}}}}%
    \put(0,0){\includegraphics[width=\unitlength,page=42]{network3.pdf}}%
    \put(0.79770631,0.43293154){\color[rgb]{0,0,0}\rotatebox{90}{\makebox(0,0)[t]{\lineheight{1.25}\smash{\begin{tabular}[t]{c}{\tiny SSC, K3S1, 96}\end{tabular}}}}}%
    \put(0,0){\includegraphics[width=\unitlength,page=43]{network3.pdf}}%
    \put(0.82034736,0.43293154){\color[rgb]{0,0,0}\rotatebox{90}{\makebox(0,0)[t]{\lineheight{1.25}\smash{\begin{tabular}[t]{c}{\tiny Deconv, K3S2, 96}\end{tabular}}}}}%
    \put(0,0){\includegraphics[width=\unitlength,page=44]{network3.pdf}}%
    \put(0.84298841,0.52559617){\color[rgb]{0,0,0}\rotatebox{90}{\makebox(0,0)[t]{\lineheight{1.25}\smash{\begin{tabular}[t]{c}{\tiny SSC, K3S1, 64}\end{tabular}}}}}%
    \put(0,0){\includegraphics[width=\unitlength,page=45]{network3.pdf}}%
    \put(0.86562946,0.52559617){\color[rgb]{0,0,0}\rotatebox{90}{\makebox(0,0)[t]{\lineheight{1.25}\smash{\begin{tabular}[t]{c}{\tiny SSC, K3S1, 64}\end{tabular}}}}}%
    \put(0,0){\includegraphics[width=\unitlength,page=46]{network3.pdf}}%
    \put(0.88827052,0.52559617){\color[rgb]{0,0,0}\rotatebox{90}{\makebox(0,0)[t]{\lineheight{1.25}\smash{\begin{tabular}[t]{c}{\tiny Deconv, K3S2, 64}\end{tabular}}}}}%
    \put(0,0){\includegraphics[width=\unitlength,page=47]{network3.pdf}}%
    \put(0.91091157,0.6182608){\color[rgb]{0,0,0}\rotatebox{90}{\makebox(0,0)[t]{\lineheight{1.25}\smash{\begin{tabular}[t]{c}{\tiny SSC, K3S1, 32}\end{tabular}}}}}%
    \put(0,0){\includegraphics[width=\unitlength,page=48]{network3.pdf}}%
    \put(0.93355262,0.6182608){\color[rgb]{0,0,0}\rotatebox{90}{\makebox(0,0)[t]{\lineheight{1.25}\smash{\begin{tabular}[t]{c}{\tiny SSC, K3S1, 32}\end{tabular}}}}}%
    \put(0,0){\includegraphics[width=\unitlength,page=49]{network3.pdf}}%
    \put(0.95619367,0.6182608){\color[rgb]{0,0,0}\rotatebox{90}{\makebox(0,0)[t]{\lineheight{1.25}\smash{\begin{tabular}[t]{c}{\tiny SSC, K3S1, 32}\end{tabular}}}}}%
    \put(0,0){\includegraphics[width=\unitlength,page=50]{network3.pdf}}%
    \put(0.05055153,0.6182608){\color[rgb]{0,0,0}\rotatebox{90}{\makebox(0,0)[t]{\lineheight{1.25}\smash{\begin{tabular}[t]{c}{\tiny Input}\end{tabular}}}}}%
    \put(0,0){\includegraphics[width=\unitlength,page=51]{network3.pdf}}%
    \put(0.89748541,0.13158999){\color[rgb]{0,0,0}\makebox(0,0)[t]{\lineheight{1.25}\smash{\begin{tabular}[t]{c}{\tiny Concatenate}\end{tabular}}}}%
    \put(0,0){\includegraphics[width=\unitlength,page=52]{network3.pdf}}%
    \put(0.89802441,0.11380043){\color[rgb]{0,0,0}\makebox(0,0)[t]{\lineheight{1.25}\smash{\begin{tabular}[t]{c}{\tiny Add}\end{tabular}}}}%
    \put(0.76816036,0.07434077){\color[rgb]{0,0,0}\makebox(0,0)[lt]{\lineheight{1.25}\smash{\begin{tabular}[t]{l}{\tiny SSC: Submanifold Sparse Convolution}\end{tabular}}}}%
    \put(0.76828527,0.05953567){\color[rgb]{0,0,0}\makebox(0,0)[lt]{\lineheight{1.25}\smash{\begin{tabular}[t]{l}{\tiny SC:  Sparse Convolution}\end{tabular}}}}%
    \put(0.7677198,0.15269428){\color[rgb]{0,0,0}\makebox(0,0)[lt]{\lineheight{1.25}\smash{\begin{tabular}[t]{l}{\tiny \textbf{Notations}}\end{tabular}}}}%
    \put(0.76756851,0.09266863){\color[rgb]{0,0,0}\makebox(0,0)[lt]{\lineheight{1.25}\smash{\begin{tabular}[t]{l}{\tiny \textbf{Abbreviations}}\end{tabular}}}}%
    \put(0.76836697,0.04562352){\color[rgb]{0,0,0}\makebox(0,0)[lt]{\lineheight{1.25}\smash{\begin{tabular}[t]{l}{\tiny K3S1(2):  Kernel size = $3\times3\times3$,}\end{tabular}}}}%
    \put(0.80674397,0.03214672){\color[rgb]{0,0,0}\makebox(0,0)[lt]{\lineheight{1.25}\smash{\begin{tabular}[t]{l}{\tiny Stride = 1(2)}\end{tabular}}}}%
  \end{picture}%
\endgroup%

			\vspace{-4pt}
			
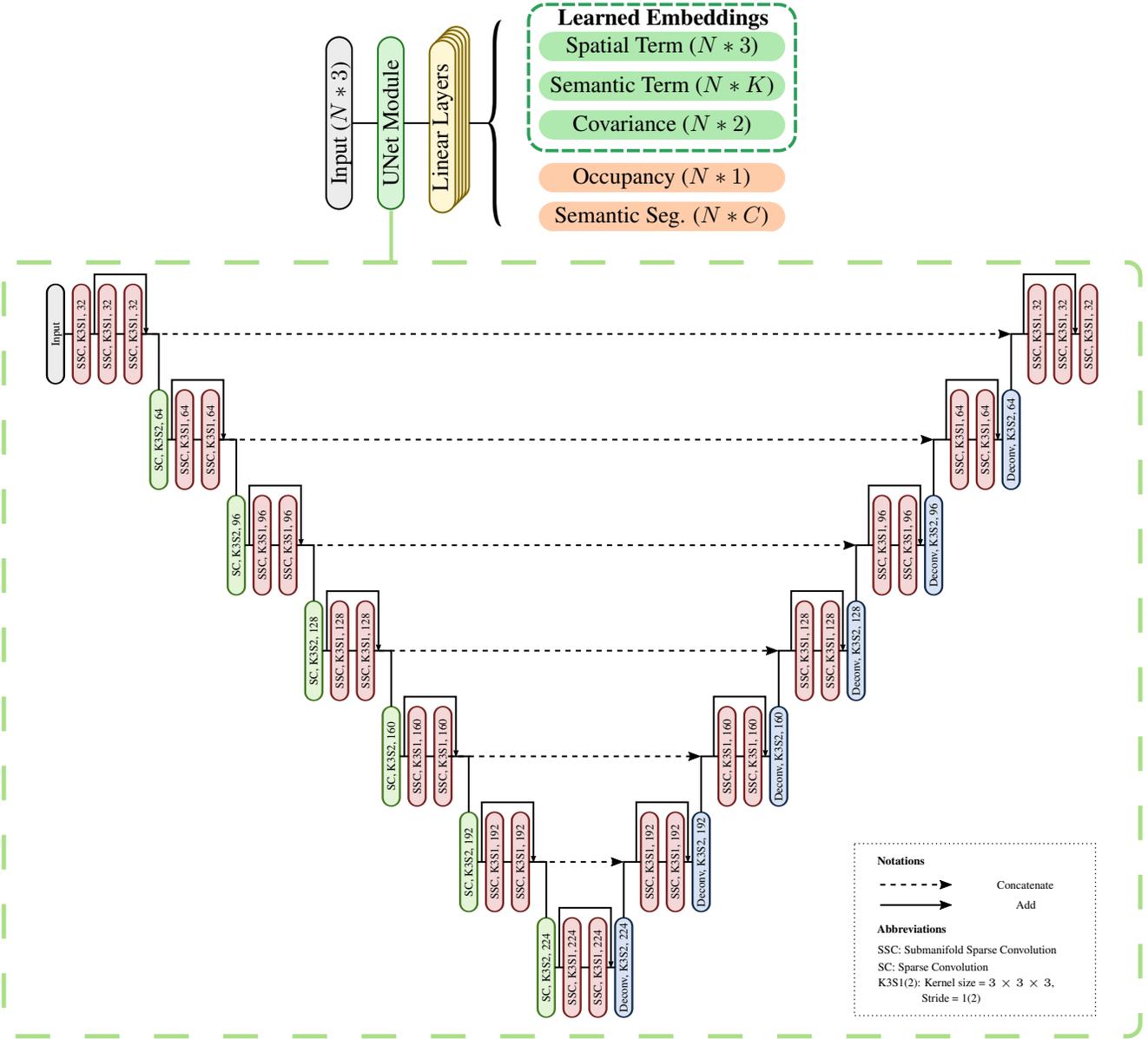
\captionof{figure}{The detailed network architecture of OccuSeg. We use a 3D UNet-like network~\cite{ronneberger2015u} as the feature extractor. Afterwards, the feature goes through different linear layers, yielding multiple learning objectives. (Best viewed on the display)}
			\label{fig:network}
		\end{center}
	}]

	\section{Network Architecture Details}
	%
	%
	
	The network architecture adopted in OccuSeg is presented in this section. We employ the widely used UNet-style network~\cite{ronneberger2015u} for feature learning as shown in Fig.~\ref{fig:network}. The network is mainly built upon submanifold sparse convolution(SSC) layers and sparse convolution layers, both of which are originally introduced by Graham~\cite{graham2014spatially}. In Fig.~\ref{fig:network}, $K$ represents the dimension of the semantic term of the learned embedding, which is set to $32$ in our experiment. Additionally, $C$ stands for the semantic class numbers, which conforms to the training dataset.
	
	\section{Additional Qualitative Comparisons}
	
	In this section, we show some more qualitative comparisons on the ScanNetV2\cite{dai2017scannet}. As shown in Fig.~\ref{fig:compare}, our results are compared with sparse convolutional networks\cite{graham20183d}, SGPN~\cite{wang2018sgpn} and multi-task metric learning method from Lahoud~\etal~\cite{lahoud20193d}. Our results generally show a better capacity of dealing with small objects, as well as produce less noise.
\twocolumn[{%
	\renewcommand\twocolumn[1][]{#1}%
    	\begin{center}
    	\centering
    	\def\svgwidth{\textwidth}
	\executeiffilenewer{inkscape/compare3.svg}{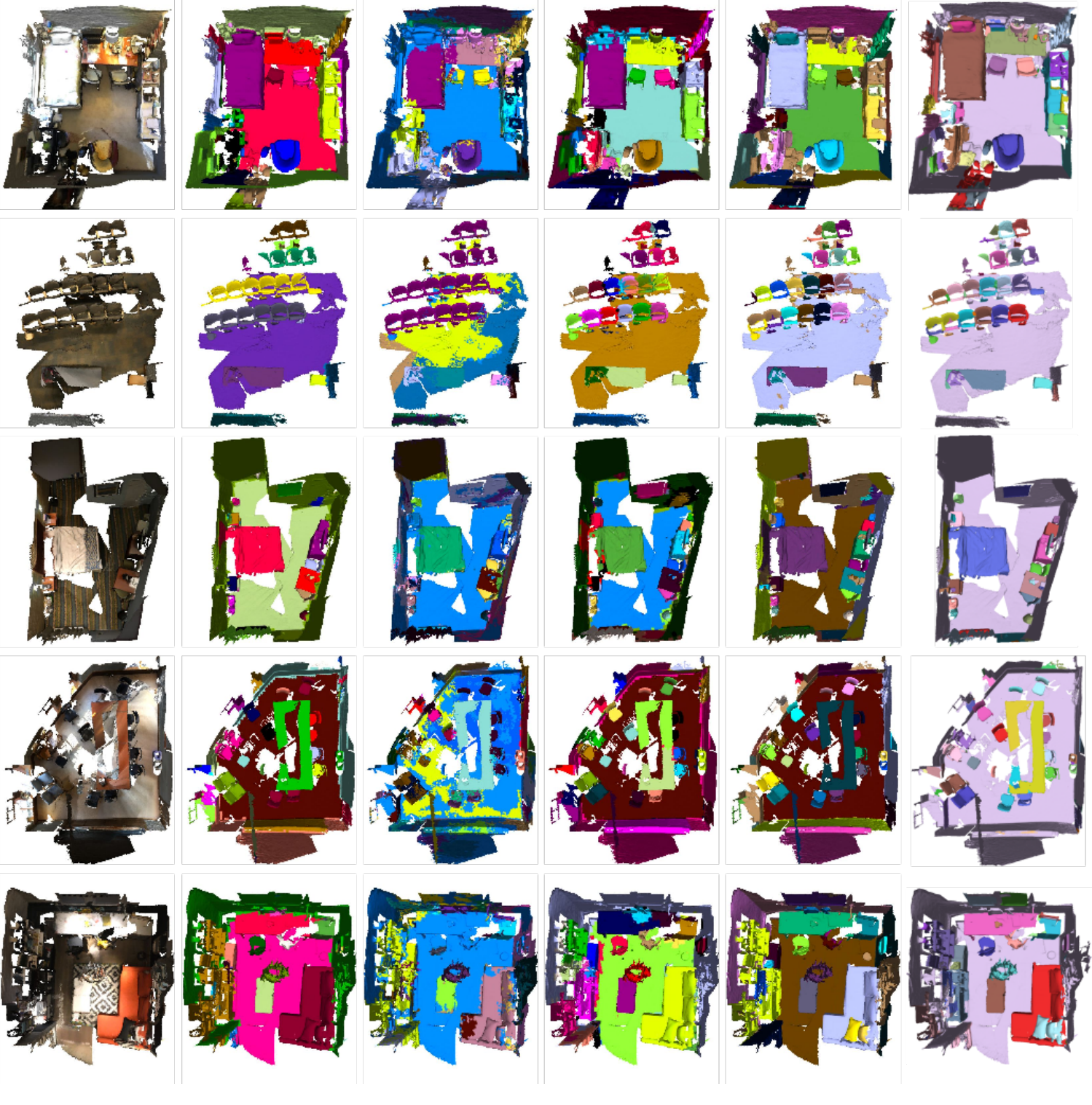}%
	{inkscape -z -D --file=inkscape/compare3.svg %
		--export-pdf=inkscape/compare3.pdf --export-latex}%
\begingroup%
  \makeatletter%
  \providecommand\color[2][]{%
    \errmessage{(Inkscape) Color is used for the text in Inkscape, but the package 'color.sty' is not loaded}%
    \renewcommand\color[2][]{}%
  }%
  \providecommand\transparent[1]{%
    \errmessage{(Inkscape) Transparency is used (non-zero) for the text in Inkscape, but the package 'transparent.sty' is not loaded}%
    \renewcommand\transparent[1]{}%
  }%
  \providecommand\rotatebox[2]{#2}%
  \newcommand*\fsize{\dimexpr\f@size pt\relax}%
  \newcommand*\lineheight[1]{\fontsize{\fsize}{#1\fsize}\selectfont}%
  \ifx\svgwidth\undefined%
    \setlength{\unitlength}{431.6657215bp}%
    \ifx\svgscale\undefined%
      \relax%
    \else%
      \setlength{\unitlength}{\unitlength * \real{\svgscale}}%
    \fi%
  \else%
    \setlength{\unitlength}{\svgwidth}%
  \fi%
  \global\let\svgwidth\undefined%
  \global\let\svgscale\undefined%
  \makeatother%
  \begin{picture}(1,1.01197708)%
    \lineheight{1}%
    \setlength\tabcolsep{0pt}%
    \put(0,0){\includegraphics[width=\unitlength,page=1]{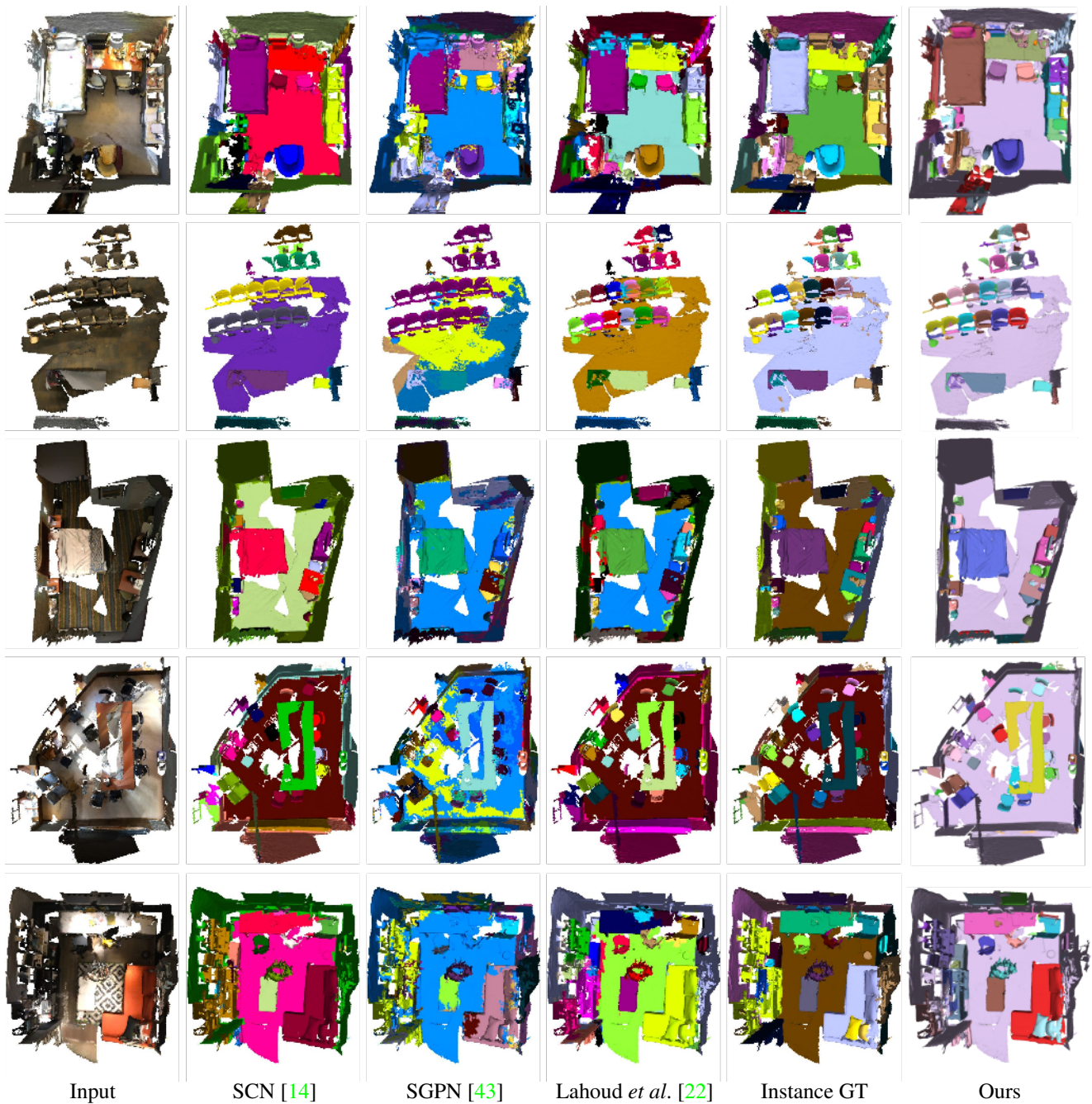}}%
    \put(0.08014853,0.00438504){\color[rgb]{0,0,0}\makebox(0,0)[t]{\lineheight{1.25}\smash{\begin{tabular}[t]{c}Input\end{tabular}}}}%
    \put(0.24785635,0.00438504){\color[rgb]{0,0,0}\makebox(0,0)[t]{\lineheight{1.25}\smash{\begin{tabular}[t]{c}SCN~\cite{graham20183d}\end{tabular}}}}%
    \put(0.41399899,0.00438504){\color[rgb]{0,0,0}\makebox(0,0)[t]{\lineheight{1.25}\smash{\begin{tabular}[t]{c}SGPN~\cite{wang2018sgpn}\end{tabular}}}}%
    \put(0.74534522,0.00438504){\color[rgb]{0,0,0}\makebox(0,0)[t]{\lineheight{1.25}\smash{\begin{tabular}[t]{c}Instance GT\end{tabular}}}}%
    \put(0.57982318,0.00438504){\color[rgb]{0,0,0}\makebox(0,0)[t]{\lineheight{1.25}\smash{\begin{tabular}[t]{c}Lahoud \etal~\cite{lahoud20193d}\end{tabular}}}}%
    \put(0.91625967,0.00438504){\color[rgb]{0,0,0}\makebox(0,0)[t]{\lineheight{1.25}\smash{\begin{tabular}[t]{c}Ours\end{tabular}}}}%
  \end{picture}%
\endgroup%

    	\vspace{2pt}
    	\captionof{figure}{More qualitative comparisons on ScanNetV2~\cite{dai2017scannet}. All the results of the previous methods are taken from Lahoud~\etal~\cite{lahoud20193d}. Note that the instance segmentation result of sparse convolution networks(SCN)~\cite{graham20183d} is obtained by showing connected components of its semantic segmentation results.}
    	\label{fig:compare}
    \end{center}
    }]

\end{document}